\documentclass{article}

\usepackage{arxiv}

\usepackage[utf8]{inputenc} 
\usepackage[T1]{fontenc}    
\usepackage{hyperref}       
\usepackage{url}            
\usepackage{booktabs}       
\usepackage{amsfonts}       
\usepackage{nicefrac}       
\usepackage{microtype}      
\usepackage{xcolor}         

\usepackage[american]{babel}

\usepackage{amsmath}
\usepackage{amssymb}
\usepackage{amsthm}
\usepackage{tikz}
\usetikzlibrary{arrows,cd,decorations.markings,shapes.geometric,shapes}
\usepackage{mathtools}
\usepackage{ebproof}

\usetikzlibrary{arrows,cd,decorations.markings,shapes.geometric,shapes}

\usepackage{tikzit}

\tikzstyle{blackDot}=[inner sep=0mm, minimum size=1.5mm, draw=black, shape=circle, draw=white, fill=black, line width=0.2mm]
\tikzstyle{boxSquare}=[fill=white, draw=black, shape=rectangle, minimum width=7mm, minimum height=5mm, font={\scriptsize}]
\tikzstyle{boxBroad}=[fill=white, draw=black, shape=rectangle, minimum width=10mm, minimum height=5mm, font={\scriptsize}]
\tikzstyle{boxSmall}=[fill=white, draw=black, shape=rectangle, minimum width=5mm, minimum height=5mm, font={\scriptsize}]
\tikzstyle{emptyText}=[fill=none, draw=none, shape=circle, font={\tiny}]
\tikzstyle{emptyTextScripsize}=[fill=none, draw=none, shape=circle, font={\scriptsize}]
\tikzstyle{state}=[fill=white, draw=black, regular polygon, regular polygon sides=3, minimum width=0.8cm, shape border rotate=180, inner sep=0pt, font={\scriptsize}]
\tikzstyle{stateLarge}=[fill=white, draw=black, regular polygon, regular polygon sides=3, minimum width=1cm, shape border rotate=180, inner sep=0pt, font={\scriptsize}]
\tikzstyle{box very broad}=[fill=white, draw=black, shape=rectangle, minimum width=20mm]
\tikzstyle{stateReverse}=[fill=white, draw=black, regular polygon, regular polygon sides=3, minimum width=0.8cm, inner sep=0pt, font={\scriptsize}]
\tikzstyle{whiteDot}=[fill=white, draw=black, shape=circle, minimum size=1.2mm, shape=circle, inner sep=0mm]

\tikzstyle{edge}=[fill=none, draw=black, line width=0.65pt, -]
\tikzstyle{white edge}=[-, draw=white, line width=1.5mm]
\tikzstyle{ddashed}=[-, draw=black, dashed]
\tikzstyle{doubleline}=[-, double, draw=black, fill=none]
\tikzstyle{redLine}=[-, draw=red]
\tikzstyle{blueLine}=[-, draw=blue]

\usepackage[all,cmtip]{xy}

\usetikzlibrary{circuits.ee.IEC}

\tikzstyle{white dot}=[inner sep=0mm, minimum size=1.5mm, draw=black, shape=circle, text depth=-0.2mm, draw=black, fill=white, tikzit category=nodes]
\tikzstyle{black dot}=[inner sep=0mm, minimum size=1.5mm, draw=black, shape=circle, draw=black, fill=black, tikzit category=nodes]
\tikzstyle{observed}=[inner sep=0mm, minimum size=5mm, draw=black, shape=circle, text depth=-0.2mm, draw=white, tikzit draw=gray, fill=white, tikzit category=dag]
\tikzstyle{latent}=[inner sep=0mm, minimum size=5mm, draw=black, shape=circle, text depth=-0.2mm, draw=black, fill=white, tikzit category=dag]
\tikzstyle{small box}=[shape=rectangle, text height=1.5ex, text depth=0.25ex, yshift=0.5mm, fill=white, draw=black, minimum height=6mm, yshift=-0.5mm, minimum width=6mm, font={\small}, tikzit category=boxes]
\tikzstyle{medium box}=[shape=rectangle, draw=black, fill=white, small box, minimum width=8mm, tikzit category=boxes]
\tikzstyle{semilarge box}=[shape=rectangle, draw=black, fill=white, small box, minimum width=12.5mm, tikzit category=boxes]
\tikzstyle{large box}=[shape=rectangle, draw=black, fill=white, small box, minimum width=15mm, tikzit category=boxes]
\tikzstyle{upground}=[circuit ee IEC, thick, ground, rotate=90, scale=1.5, inner sep=-2mm, tikzit shape=circle, tikzit fill=blue, tikzit category=points]
\tikzstyle{downground}=[circuit ee IEC, thick, ground, rotate=-90, scale=1.5, inner sep=-2mm, tikzit shape=circle, tikzit fill=green, tikzit category=points]
\tikzstyle{point}=[regular polygon, regular polygon sides=3, draw, scale=0.75, inner sep=-0.5pt, minimum width=9mm, fill=white, regular polygon rotate=180, tikzit category=points]
\tikzstyle{copoint}=[regular polygon, regular polygon sides=3, draw, scale=0.75, inner sep=-0.5pt, minimum width=9mm, fill=white, tikzit category=points]
\tikzstyle{uniform}=[point, fill=gray, tikzit shape=circle, scale=0.5]
\tikzstyle{label}=[font={\footnotesize}, text height=1.5ex, text depth=0.25ex, tikzit draw=blue, tikzit fill=white, tikzit category=labels]
\tikzstyle{left label}=[label, anchor=east, xshift=2mm, tikzit draw=green, tikzit fill=white, tikzit category=labels]
\tikzstyle{right label}=[label, anchor=west, xshift=-2mm, tikzit draw=purple, tikzit fill=white, tikzit category=labels]
\tikzstyle{disintegration}=[draw=black, fill={gray!50}, tikzit fill=gray, shape=rectangle, minimum width=1.6cm, minimum height=1.2cm, opacity=0.3]
\tikzstyle{empty diag}=[shape=rectangle, draw=darkgray, dashed, minimum width=8mm, minimum height=8mm, yshift=0.5mm]

\tikzstyle{diredge}=[->, >=latex]
\tikzstyle{dashed edge}=[-, dashed]

\pgfdeclarelayer{edgelayer}
\pgfdeclarelayer{nodelayer}
\pgfsetlayers{edgelayer,nodelayer,main}
\tikzstyle{none}=[]

\tikzstyle{morphism}=[fill=white, draw=black, shape=rectangle]
\tikzstyle{medium box}=[fill=white, draw=black, shape=rectangle, minimum width=0.8cm, minimum height=0.9cm]
\tikzstyle{large morphism}=[fill=white, draw=black, shape=rectangle, minimum width=1.7cm, minimum height=1cm]
\tikzstyle{bn}=[fill=black, draw=black, shape=circle, inner sep=1.5pt]
\tikzstyle{state}=[fill=white, draw=black, regular polygon, regular polygon sides=3, minimum width=0.8cm, shape border rotate=180, inner sep=0pt]
\tikzstyle{medium state}=[fill=white, draw=black, regular polygon, regular polygon sides=3, minimum width=1.3cm, inner sep=0pt, shape border rotate=180]
\tikzstyle{large state}=[fill=white, draw=black, regular polygon, regular polygon sides=3, minimum width=2.2cm, shape border rotate=180, inner sep=0pt]
\tikzstyle{wn}=[fill=white, draw=black, shape=circle, inner sep=1.5pt]

\tikzstyle{arrow}=[->]
\tikzstyle{dashed box}=[-, dashed]

\tikzset{baseline=(current  bounding  box.center)}
\tikzset{every picture/.append style={scale=0.5}}

\usepackage{tikzit}

\usetikzlibrary{circuits.ee.IEC}

\usetikzlibrary{calc,arrows.meta}

\newcommand\cofib\rightarrowtail

\newcommand\mdel[1]{}

\usepackage{txfonts}
\usepackage{pxfonts}

\makeatletter
\newcommand{\xdashrightarrow}[2][]{\ext@arrow 0359\rightarrowfill@@{#1}{#2}}
\newcommand*{\doublerightarrow}[2]{\mathrel{
  \settowidth{\@tempdima}{$\scriptstyle#1$}
  \settowidth{\@tempdimb}{$\scriptstyle#2$}
  \ifdim\@tempdimb>\@tempdima \@tempdima=\@tempdimb\fi
  \mathop{\vcenter{
    \offinterlineskip\ialign{\hbox to\dimexpr\@tempdima+1em{##}\cr
    \rightarrowfill\cr\noalign{\kern.5ex}
    \rightarrowfill\cr}}}\limits^{\!#1}_{\!#2}}}
\newcommand*{\triplerightarrow}[1]{\mathrel{
  \settowidth{\@tempdima}{$\scriptstyle#1$}
  \mathop{\vcenter{
    \offinterlineskip\ialign{\hbox to\dimexpr\@tempdima+1em{##}\cr
    \rightarrowfill\cr\noalign{\kern.5ex}
    \rightarrowfill\cr\noalign{\kern.5ex}
    \rightarrowfill\cr}}}\limits^{\!#1}}}
\makeatother

\makeatletter
\newcommand{\twoarrows}[3][0.2ex]{%
  \mathrel{\mathpalette\twoarrows@{{#1}{#2}{#3}}}%
}
\newcommand{\twoarrows@}[2]{\twoarrows@@#1#2}
\newcommand{\twoarrows@@}[4]{%
  \vcenter{\offinterlineskip\m@th
    \ialign{\hfil##\hfil\cr
      $#1#3$\cr
      \noalign{\vskip#2}
      $#1#4$\cr
    }%
  }%
}
\makeatother

\newcommand{\beq}{\begin{equation}}
\newcommand{\eeq}{\end{equation}}

\newcommand{\cop}{\mathsf{copy}}
\newcommand{\del}{\mathsf{del}}

\newtheorem{theorem}{Theorem}
\newtheorem{definition}{Definition}
\newtheorem{lemma}{Lemma}
\newtheorem{example}{Example}

\usepackage[boxruled]{algorithm2e}
\usepackage{algorithmic}

\usepackage{multirow}
\usepackage{booktabs}
\usepackage{balance}
\usepackage{subfig}

\DeclareFontFamily{U}{dmjhira}{}
\DeclareFontShape{U}{dmjhira}{m}{n}{ <-> dmjhira }{}

\DeclareRobustCommand{\yo}{\text{\usefont{U}{dmjhira}{m}{n}\symbol{"48}}}

\usepackage{natbib} 
\usepackage{mathtools} 
\usepackage{booktabs} 
\usepackage{tikz} 

\title{Topos Causal Models\thanks{Draft under submission.} }

\author{ Sridhar Mahadevan \\
	Adobe Research and University of Massachusetts, Amherst\\
	\texttt{smahadev@adobe.com, mahadeva@umass.edu}
}

\hypersetup{
pdftitle={A template for the arxiv style},
pdfsubject={q-bio.NC, q-bio.QM},
pdfauthor={David S.~Hippocampus, Elias D.~Striatum},
pdfkeywords={First keyword, Second keyword, More},
}

\begin{document}
\maketitle

\begin{abstract}
We propose topos causal models (TCMs), a novel class of causal models that exploit the key properties of a topos category: they are (co)complete, meaning  all (co)limits exist, they admit a {\em subobject classifier}, and allow {\em exponential objects}. The main goal of this paper is to show that these properties are central to many applications in causal inference. For example, subobject classifiers allow a categorical formulation of causal intervention, which creates sub-models. Limits and colimits allow causal diagrams of arbitrary complexity to be ``solved", using a novel interpretation of causal approximation. Exponential objects enable reasoning about equivalence classes of operations on causal models, such as covered edge reversal and causal homotopy. Analogous to structural causal models (SCMs), TCMs are defined by a collection of functions, each defining a ``local autonomous" causal mechanism that assemble to induce a unique global function from exogenous to endogenous variables. Since the category of TCMs is (co)complete, which we prove in this paper, every causal diagram has a ``solution" in the form of a (co)limit: this implies that any arbitrary causal  model can be ``approximated" by some global function with respect to the morphisms going into or out of the diagram. Natural transformations are crucial in measuring the quality of approximation. In addition, we show that causal interventions are modeled by subobject classifiers: any sub-model is defined by a monic arrow into its parent model. Exponential objects permit reasoning about entire classes of causal equivalences and interventions. Finally, as TCMs form a topos, they admit an internal logic defined as a Mitchell-Benabou language with an associated Kripke-Joyal semantics. We show how to reason about causal models in TCMs using this internal logic.  
\end{abstract}


\newpage 

\tableofcontents

\section{Introduction}\label{sec:intro}

In recent years, there has been significant interest in categorical models of causality, based on symmetric monoidal categories \citep{fong:ms,fritz:jmlr,Cho_2019,string-diagram-surgery}, as well as simplicial sets and higher-order categories \citep{DBLP:journals/entropy/Mahadevan23}. 
\citep{DBLP:journals/entropy/Mahadevan23,cktheory} introduced the framework of {\em universal causality} based on the notion of universal properties in category theory \citep{riehl2017category}: a causal property is universal if it can be defined in terms of an {\em initial} or {\em final} object in a category of causal diagrams, or in terms of a {\em causal representable functor} using the Yoneda Lemma. For example, a structural causal model (SCM) \citep{pearl-book} is ostensibly defined as a (deterministic) mapping from a collection of exogenous variables into a collection of endogenous variables, derived by ``collating" local functions that serve as independent causal mechanisms \citep{scm-lewis,icm}. However, SCMs can be further analyzed in terms of their universal properties, such as categorical product, coproduct, limits and colimits, equalizers and coequalizers etc. These latter properties can be shown formally to be initial or final objects in a category of diagrams \citep{riehl2017category}, or as representable functors through the Yoneda Lemma, and finally also characterized in terms of the universal properties of {\em Kan extensions} \citep{maclane:71}. In summary, categorical approaches fundamentally differ from past work in causality in their focus on the elucidation of universal properties. 

Markov categories \citep{Fritz_2020} define a broad unifying framework for probabilistic inference and statistics using an elegant string diagrammatic language \citep{Selinger_2010}, which allows rigorous proofs of classical results and yet is significantly simpler than   the traditional ``assembly language" formalism of measure theory \citep{halmos:book}. Any causal model based on graphs  \citep{pearl-book,hedge,spirtes:book} or other algebraic formalisms, such as integer-valued multisets \citep{studeny2010probabilistic},  can be translated into a string diagram over a symmetric monoidal category, or a simplicial set.  Operations on causal models, such as interventions, can be modeled as functors on the objects of the associated symmetric monoidal category or simplicial set. Categorical approaches to causality also extend to the {\em potential outcomes} counterfactual framework \citep{rubin-book}. 

Our main contribution in this paper is present a {\em topos-theoretic} view of causality. A topos is a type of category \citep{maclane:71}, which is particularly well-suited to modeling operations that are ``set-like" \citep{maclane:sheaves}. It also features an internal logical language \citep{goldblatt:topos}. We claim that a topos provides three universal properties  that make it natural as a category to do causal inference in: it provides a general theory for how to combine local functions, which can be viewed as ``independent causal mechanisms" \citep{icm}, into a consistent global function building on the theory of sheaves in a topos \citep{maclane1992sheaves}. It enables a generic way to define causal interventions using a subobject classifier in a topos  category \citep{Johnstone:topostheory}. Finally, it gives an internal logical language for causal and counterfactual reasoning \citep{bell}. We illustrate the application of this topos-theoretic view of causality both to structural causal models (SCMs), as well as previous categorical models of causality, and to Lewis' theory of counterfactuals \citep{DBLP:journals/jphil/Lewis73}.

In addition, a topos causal model introduces into the causal inference literature new ways of combining causal models, using universal constructions such as {\em limits}, {\em colimits}, {\em (co)equalizers}, and their specializations such as {\em pullbacks} and {\em pushforwards}. These universal constructions, whose role in causal inference was first articulated in \citep{DBLP:journals/entropy/Mahadevan23}, show that it is possible to ``solve" causal diagrams of arbitrary complexity using a novel interpretation of what an approximation is. More formally, a causal diagram $F: {\cal J} \rightarrow {\cal C_T}$  is defined by some functor $F$ from an indexing category ${\cal J}$ into some TCM category ${\cal C_T}$. We can exploit the topos-like nature of the TCM category to show that all causal diagrams have a solution in terms of their limits or colimits. 

At the outset, it is important to stress that our paper is not intended to be an algorithmic contribution to the vast literature on causal discovery (for a recent survey of the past $30$ years of causal discovery, see \citep{zanga2023surveycausaldiscoverytheory}), or indeed a  proposal for causal inference on particular types of data-generating distributions \citep{do-finetti}. Rather, in the spirit of categorical characterizations, we aim to uncover more general principles that apply to a broad class of algorithms and models. For example, {\em exchangeable} causal models have recently been proposed as a way to instantiate the independent causal mechanism hypothesis \citep{icm,do-finetti}. The fundamental classical result of de Finetti on exchangeability can be elegantly generalized using Markov categories \citep{fritz_de_finetii}, which can potentially be combined with our topos-theoretic view to yield a deeper understanding of  discovering causal structures from exchangeable processes. 


\section{High-Level Overview of the Framework}

\begin{figure}[t]
    \centering
    \includegraphics[width=0.75\linewidth]{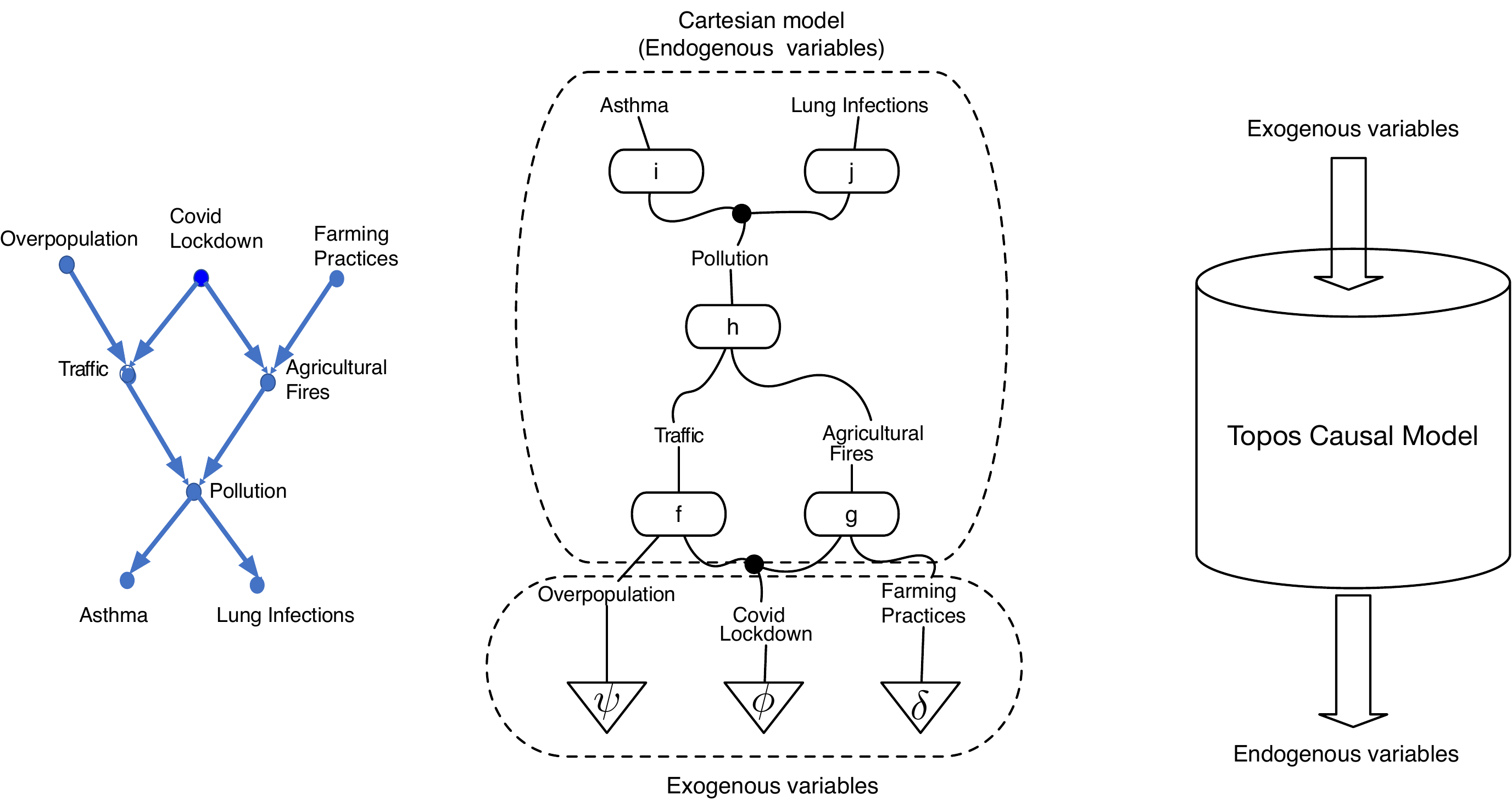}
    \caption{Left: a topos causal model (SCM) of pollution in New Delhi, India; Middle: A TCM encoded as a string diagram in a Markov category \citep{Fritz_2020}. Right: Each object in a TCM is a causal model, which can be thought of  as a ``blackbox" function mapping exogenous variables to endogenous variables.}
    \label{fig:tcm-example}
\end{figure}

Before getting into the technical details, we give a high level overview of the TCM framework in this section. TCMs introduce quite a few novel notions into the literature of causal inference, and it will take some space to unpack these ideas. We include a brief review of basic concepts in category theory in Section~\ref{introcat} at the end of the paper, which the reader is encouraged to consult as needed. 

Figure~\ref{fig:tcm-example} illustrates the three ways to understand the TCM framework. It can be viewed as a way to model traditional causal graphical models \citep{pearl:bnets-book,spirtes:book}, defined by directed acyclic graph (DAG) models or some more elaborate type of graph \citep{hedge}. It is worth immediately noting that all directed graphs form a topos functor category \citep{vigna2003guidedtourtoposgraphs}, leading immediately to a novel way to analyze causal graphical models. TCMs can be thought of in terms of {\em string diagrams} in a {\em Markov category} \citep{Fritz_2020}, with the important caveat that we are primarily only interested in Markov categories that form toposes. Finally, we can abstract out a TCM object as simply a ``blackbox" function that maps some collection of exogenous variables (e.g., ``overpopulation", or "Farming Practices" into some set of endogenous variables, e.g., ``Asthma" or ``Pollution"). 

\begin{enumerate}
    \item {\em Causal models are diagrams}: As introduced in \citep{DBLP:journals/entropy/Mahadevan23}, in TCMs, causal models are {\em diagrams}, or formally functors $F: {\cal J} \rightarrow {\cal C_T}$, where ${\cal J}$ is a ``small" category that represents the structure of the model, and ${\cal C_T}$ specifies the causal model itself. Canonical diagrams for TCMs include {\em pullbacks} $\bullet \rightarrow \bullet \leftarrow \bullet$, {\em pushforwards} $\bullet \rightarrow \bullet \rightarrow \bullet$, {\em equalizers}, which take the abstract form $\bullet \rightarrow \bullet \twoarrows[0.4ex]{\rightarrow}{\rightarrow}  \bullet$, and {\em co-equalizers}, which take the form $\bullet \twoarrows[0.4ex]{\rightarrow}{\rightarrow}  \bullet \rightarrow \bullet$. These abstract diagrams are mapped into actual causal relationships in the category ${\cal C}_{T}$ by some functor ${\cal J}$.

    \item {\em Limits and colimits of causal model}: Another novel idea in this paper is to determine the limit or colimit \citep{maclane:71} of a causal model. Since TCMs are defined in a topos, all (finite) limits and colimits exist. We will show that these limits and colimits essentially give us a way to ``approximate" an arbitrary TCM model. Applied to SCMs, for example, since these are defined as deterministic functions over sets, we can use the property that the category ${\cal C_{\bf Sets}}$ is (co)complete to determine the (co)limit of any SCM. 

    \item {\em Causal interventions are modeled using sub-object classifiers}: A key property of a topos \citep{Johnstone:592033} is that it possesses a sub-object classifier. In the category ${\cal C_{\bf Set}}$, the sub-object classifier derives from the property that a subset $A \subset B$ induces a Boolean characteristic function $\psi_B$ that takes the value $\psi_B(x) = 1$ for all elements $x \in B$ that are also in $A$. For an arbitrary TCM object $c \in {\cal C_T}$, a sub-object classifier is defined through {\em monic} (injective) arrows $c \rightarrowtail d$, which are all constructed as a pullback of the primary monic arrow ${\bf 1} \rightarrow \Omega$, where ${\bf 1}$ is the terminal element of the topos category ${\cal C_T}$ and $\Omega$ is an object that captures the non-Boolean ``degrees of truth" in a general topos. 
\end{enumerate}

\section{Topos Causal Models} 

We now introduce topos causal models (TCMs) more formally. Suitable background material for category theory and topos theory is given in the appendices. 

\begin{definition}
\label{tcmcat}
The category ${\cal C}_{TCM}$ of topos causal models is defined as a collection of objects, each of which defines a triple $\langle U, V, F \rangle$ where $V = \{V_1, \ldots, V_n \}$ is  a set of {\em endogenous} variables, $U$ is a set of {\em exogenous} variables, $F$ is a function from $F$ from  $U$ to $V$.  A submodel $M' = \langle U', V', F' \rangle $ of $M$ is any subobject of $M$.  The  effect of an action on $M$ is given by some submodel $M'$. Finally, let  $Y$ be a variable in $V$, and let $X$ be a subset of $V$. The potential outcome in response to an action that generates a submodel $M'$ is the solution of $Y$ in the submodel $M'$. 
\end{definition}

The set of arrows or morphisms between two objects $c$ and $d$ in the category ${\cal C}_{TCM}$, denoted ${\cal C}_{TCM}(c,d)$, represent ways of transitioning from TCM object $c$ to $d$. For example, if $d$ is a submodel of $c$, then the arrow defines a {\bf do} causal intervention. In reasoning about causal equivalences, it will be important to represent morphisms like {\em covered edge reversals} \citep{meek2013causalinferencecausalexplanation,chickering:jmlr}, which generate categorical structures called {\em groupoids} \citep{may1999concise}. Our goal is to construct homotopies between different TCMs, treated as objects in a model category \citep{Quillen:1967}. The concept of {\em homotopy} originally arose in algebraic topology as a way to characterize when two topological spaces were essentially the same, such as a coffee cup is like a doughnut since they both have one hole, and they both are topologically distinct from a ``wedge sum of two circles" that looks like  a figure eight. 

\subsection{SCMs as a Topos Causal Model}

From Definition~\ref{tcmcat}, it should be fairly clear that structural causal models (SCMs) are obviously a special case of TCMs, where interventions are specified by clamping a set of variables $X=x$. For completeness, we define a category ${\cal C}_{SCM}$ whose objects are indeed SCMs.

\begin{definition}
\label{scmcat}
The category ${\cal C}_{SCM}$ of structural causal models is defined as a collection of objects, each of which is a triple $\langle U, V, F \rangle$ where $V = \{V_1, \ldots, V_n \}$ is  a set of {\em endogenous} variables, $U$ is a set of {\em exogenous} variables, $F$ is a set $\{f_1, \ldots, f_n \}$ of ``local functions" $f_i: U \cup (V \setminus V_i) \rightarrow V_i$ whose composition induces  a unique function $F$ from  $U$ to $V$.  Let $X$ be a subset of variables in $V$, and $x$ be a particular realization of $X$.  A submodel $M_x = \langle U, V, F_x \rangle $ of $M$ is the causal model $ M_x =  \langle U, V, F_x \rangle$, where $F_x = \{f_i : V_i \notin X \} \cup \{X = x \}$. The  effect of an action $\mbox{do}(X=x)$ on $M$ is given by the submodel $M_x$. Finally, let  $Y$ be a variable in $V$, and let $X$ be a subset of $V$. The potential outcome of $Y$ in response to an action $\mbox{do}(X=x)$, denoted $Y_x(u)$, is the solution of $Y$ for the set of equations $F_x$. 
\end{definition}

The set of arrows or morphisms between two objects $c$ and $d$ in the category ${\cal C}_{SCM}$, denoted ${\cal C}_{SCM}(c,d)$, represent ways of transitioning from SCM object $c$ to $d$. For example, if $d$ is a submodel of $c$, then the arrow defines a {\bf do} causal intervention. In reasoning about causal equivalences, it will be important to represent morphisms like {\em covered edge reversals} \citep{meek2013causalinferencecausalexplanation,chickering:jmlr}, which generate categorical structures called {\em groupoids} \citep{may1999concise}. Our goal is to construct homotopies between different SCMs, treated as objects in a model category \citep{Quillen:1967}. The concept of {\em homotopy} originally arose in algebraic topology as a way to characterize when two topological spaces were essentially the same, such as a coffee cup is like a doughnut since they both have one hole, and they both are topologically distinct from a ``wedge sum of two circles" that looks like  a figure eight. 


\section{Topos Causal Models are (co)Complete}
 
 To prove that the category ${\cal C}_{TCM}$ (or its special case ${\cal C}_{SCM}$) defines a topos, we need to first show that it is (co)complete (see Theorem~\ref{scmcocomplete} below). (Co)limits are generally modeled as {\em diagrams} $F: {\cal J} \rightarrow {\cal C}$, where $F$ defines a functor that defines a diagram over category ${\cal C}$ of shape ${\cal J}$. To explain these concepts better, let us take some simple examples from the category of sets. 

\begin{example} 
If  we consider a small  ``discrete'' category ${\cal D}$ whose only morphisms are identity arrows, then the colimit of a functor ${\cal F}: {\cal D} \rightarrow {\cal C}$, defined by the diagram $\bullet \bullet \cdots \bullet$ is the {\em categorical coproduct} of ${\cal F}(D)$ for $D$, an object of category {\cal D}, is denoted as 
\[ \mbox{Colimit}_{\cal D} F = \bigsqcup_D {\cal F}(D) \]

In the special case when the category {\cal C} is the category ${\cal C}_{\bf Set}$, then the colimit of this functor is simply the disjoint union of all the sets $F(D)$ that are mapped from objects $D \in {\cal D}$. Note that each $\bullet$ is mapped by the functor $F$ to an actual set. 
\end{example} 

\begin{example} 
Dual to the notion of colimit of a functor is the notion of {\em limit}. Once again, if we consider a small  ``discrete'' category ${\cal D}$ whose only morphisms are identity arrows, then the limit of a functor ${\cal F}: {\cal D} \rightarrow {\cal C}$ is the {\em categorical product} of ${\cal F}(D)$ for $D$, an object of category {\cal D}, is denoted as 
\[ \mbox{limit}_{\cal D} F = \prod_D {\cal F}(D) \]

In the special case when the category {\cal C} is the category ${\cal C}_{\bf Set}$, then the limit of this functor is simply the Cartesian product of all the sets $F(D)$ that are mapped from objects $D \in {\cal D}$. The diagram remains the same $\bullet \bullet \cdots \bullet$. 
\end{example} 

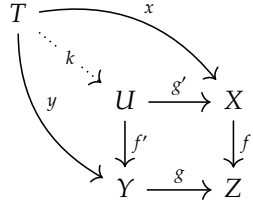
\begin{figure}[h]
\centering
\begin{tikzcd}
  T
  \arrow[drr, bend left, "x"]
  \arrow[ddr, bend right, "y"]
  \arrow[dr, dotted, "k" description] & & \\
    & 
    U\arrow[r, "g'"] \arrow[d, "f'"]
      & X \arrow[d, "f"] \\
& Y \arrow[r, "g"] &Z
\end{tikzcd}
\caption{
 Universal property of pullback mappings. } 
\label{pullback}
\end{figure}

Figure~\ref{pullback}  illustrates the limit of a more complex diagram referred as a {\em {pullback}}, whose diagram is written abstractly as $\bullet \rightarrow \bullet \leftarrow \bullet$.  Note in Figure~\ref{pullback}, the functor maps the diagram $\bullet \rightarrow \bullet \leftarrow \bullet$ to actual objects in the category $Y \xrightarrow[]{g} Z \xleftarrow[]{f} X$. The universal property of the pullback square with the objects $U,X, Y$ and $Z$ implies that the composite mappings $g \circ f'$ must equal $g' \circ f$. In this example, the morphisms $f$ and $g$ represent a {\em {pullback}} pair, as they share a common co-domain $Z$. The pair of morphisms $f', g'$ emanating from $U$ define a {\em {cone}}, because the pullback square ``commutes'' appropriately. Thus, the pullback of the pair of morphisms $f, g$ with the common co-domain $Z$ is the pair of morphisms $f', g'$ with common domain $U$. Furthermore, to satisfy the universal property, given another pair of morphisms $x, y$ with common domain $T$, there must exist another morphism $k: T \rightarrow U$ that ``factorizes'' $x, y$ appropriately, so that the composite morphisms $f' \ k = y$ and $g' \ k = x$. Here, $T$ and $U$ are referred to as {\em cones}, where $U$ is the limit of the set of all cones ``above'' $Z$. If we reverse arrow directions appropriately, we get the corresponding notion of pushforward. 

Other common diagrams are {\em equalizers}, which take the abstract form $\bullet \rightarrow \bullet \twoarrows[0.4ex]{\rightarrow}{\rightarrow}  \bullet$, and {\em co-equalizers}, which take the form $\bullet \twoarrows[0.4ex]{\rightarrow}{\rightarrow}  \bullet \rightarrow \bullet$. Once again, these abstract diagrams are mapped into actual relationships in the category ${\cal C}_{SCM}$ by some functor. An example of an equalizer diagram from the category ${\cal C}_{\bf Set}$ gives intuition: 
\begin{example}
The solution to the {\em equalizer} diagram $\bullet \rightarrow \bullet \twoarrows[0.4ex]{\rightarrow}{\rightarrow}  \bullet$ in the category of sets ${\cal C}_{\bf Set}$ is given by the the set $\{x \in X | f(x) = g(x) \}$, where the diagram functor $F: {\cal J} \rightarrow {\cal C}_{\bf Set}$ maps the two parallel arrows to  ${\bf 1}  \xrightarrow{x} X \twoarrows[0.2ex]{\xrightarrow{f}}{\xrightarrow{g}}  Y$, where ${\bf 1} \xrightarrow{x} X$ is an element selector morphism. 
\end{example}
For any object $c \in C$ and any category $J$, the {\em constant functor} $c: J \rightarrow C$ maps every object $j$ of $J$ to $c$ and every morphism $f$ in $J$ to the identity morphisms $1_c$. We can define a constant functor embedding as the collection of constant functors $\Delta: C \rightarrow C^J$ that send each object $c$ in $C$ to the constant functor at $c$ and each morphism $f: c \rightarrow c'$ to the constant natural transformation, that is, the natural transformation whose every component is defined to be the morphism $f$. 

\begin{definition}
    A {\bf cone over} a diagram $F: J \rightarrow C$ with the {\bf summit} or {\bf apex} $c \in C$ is a natural transformation $\lambda: c \Rightarrow F$ whose domain is the constant functor at $c$. The components $(\lambda_j: c \rightarrow Fj)_{j \in J}$ of the natural transformation can be viewed as its {\bf legs}. Dually, a {\bf cone under} $F$ with {\bf nadir} $c$ is a natural transformation $\lambda: F \Rightarrow c$ whose legs are the components $(\lambda_j: F_j \rightarrow c)_{j \in J}$.

\[\begin{tikzcd}
	&& c \\
	\\
	{F j} &&&& Fk
	\arrow["{\lambda_j}", from=1-3, to=3-1]
	\arrow["{\lambda_k}"', from=1-3, to=3-5]
	\arrow["{F f}", from=3-1, to=3-5]
\end{tikzcd}\]
    
\end{definition}
Cones under a diagram are referred to usually as {\em cocones}. Using the concept of cones and cocones, we can now formally define the concept of limits and colimits more precisely. 
\begin{definition}
    For any diagram $F: J \rightarrow C$, there is a functor $\mbox{Cone}(-, F): C^{op} \rightarrow \mbox{{\bf Set}}$, which sends $c \in C$ to the set of cones over $F$ with apex $c$. Using the Yoneda Lemma, a {\bf limit} of $F$ is defined as an object $\lim F \in C$ together with a natural transformation $\lambda: \lim F \rightarrow F$, which can be called the {\bf universal cone} defining the natural isomorphism $C(-, \lim F) \simeq \mbox{Cone}(-, F)$.  Dually, for colimits, we can define a functor $\mbox{Cone}(F, -): C \rightarrow \mbox{{\bf Set}}$ that maps object $c \in C$ to the set of cones under $F$ with nadir $c$. A {\bf colimit} of $F$ is a representation for $\mbox{Cone}(F, -)$. Once again, using the Yoneda Lemma, a colimit is defined by an object $\mbox{Colim} F \in C$ together with a natural transformation $\lambda: F \rightarrow \mbox{colim} F$, which defines the {\bf colimit cone} as the natural isomorphism $C(\mbox{colim} F, -) \simeq \mbox{Cone}(F, -)$. 
\end{definition}
Limit and colimits of diagrams over arbitrary categories can often be reduced to the case of their corresponding diagram properties over sets.
\begin{figure}
    \centering
    \includegraphics[scale=0.4]{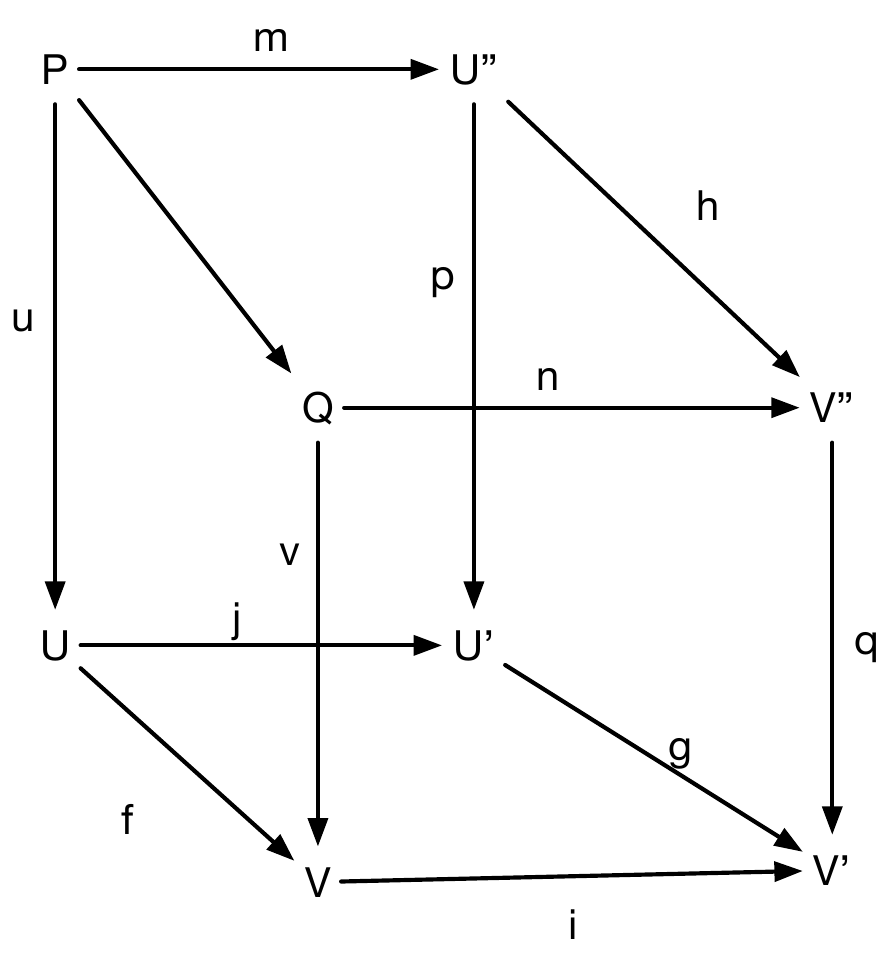}
    \caption{Figure showing that ${\cal C}_{TCM}$ has all limits and colimits.}
    \label{pullback-gdc}
\end{figure}
 Note that by Definition~\ref{scmcat}, an SCM comprises of an indexed collection of ``local" set functions $f_i: U \cup (V \setminus V_i) \rightarrow V_i$. Given two SCM objects $m$ and $m'$, we can compute (co)limits over each pair of corresponding function $f_i$ and $f'_i$ individually, and because these are set-valued functions, they must exist. 
\begin{theorem}
\label{scmcocomplete}
    The  category ${\cal C}_{TCM}$, and its special case ${\cal C}_{TCM}$, is {\em (co) complete}. 
\end{theorem}

{\bf Proof of Theorem~\ref{scmcocomplete}:} We will prove  Theorem ~\ref{scmcocomplete} showing that the category ${\cal C}_{TCM}$ has all limits and colimits. Formally, this result requires showing that all pullback diagrams of the form $\bullet \rightarrow \bullet \leftarrow \bullet$, and all pushout diagrams of the form $\bullet \leftarrow \bullet \rightarrow \bullet$ have a solution, as well as (co)equalizer diagrams of the form $\bullet \rightarrow \bullet \twoarrows[0.4ex]{\rightarrow}{\rightarrow}  \bullet$ and $\bullet \twoarrows[0.4ex]{\rightarrow}{\rightarrow}  \bullet \rightarrow \bullet$, respectively, exist, all of which build on the property that in the category of sets ${\cal C}_{\bf Set}$, these diagrams are solvable. For brevity, we will just illustrate the argument for pullback diagrams. Consider the cube shown in Figure~\ref{pullback-gdc}. Here, $f$, $g$, and $h$ are the unique functions defining three TCMs, each mapping some exogenous variables -- U, U', and U", respectively -- to some endogenous variables -- V, V', and V", respectively.  The arrow $i$ between endogenous variables V to V', and arrow $j$ from endogenous variables U to U' ensure that the bottom face of the cube is a commutative diagram, and the arrow $p$, from exogenous variables U" to exogenous variables U', and arrow $q$ from endogenous variables V" to V',  ensure the right face of the cube is a commutative diagram. The arrow from $P$ to $Q$ exists because looking at the front face of the cube, $Q$ is the pullback of $i$ and $q$, which must exist because we are in the category of sets ${\cal C}_{\bf Set}$, which has all pullbacks. Similarly, the back face of the cube is a pullback of $j$ and $p$, which is again a pullback in ${\cal C}_{\bf Set}$. Summarizing, $\langle u, v \rangle$ and $\langle m, n \rangle$ are the pullbacks of $\langle i, j \rangle$ and $\langle p, q \rangle$. The proof that ${\cal C}_{SCM}$ has all pushouts (limits) is similar. $\qed$

\section{TCMs form a Topos} 

We now show that in addition to being (co)complete, the category ${\cal C}_{TCM}$ (and it special case, ${\cal C}_{SCM}$) does form a topos. 
Formulating causal inference using TCMs results in three key properties (i) a categorical framework for ``collating"  local functions, which in SCMs \citep{scm-lewis} and statistical approaches \citep{icm} are intended to define independent causal mechanisms, into a unique globally consistent function using the universal property of  sheaves on a topos \citep{maclane1992sheaves}; (ii) a generic way to define causal interventions inducing submodels using the universal properties of  {\em subobject classifiers} in a topos category ${\cal C}_{TCM}$; and finally, (iii) a formal internal intuitionistic language for causal and counterfactual reasoning that deeply reflects the abstract topological structure of a topos. This internal language provides a new way to define causal reasoning over a typed logic \citep{Johnstone:592033}, or a local set theory \citep{bell}. 

\subsection{Brief Introduction to Topos Theory} 

A topos \citep{Johnstone:topostheory} is a ``set-like" category which generalizes all common operations on sets (see Table~\ref{setvscategories}). Thus, the generalization  of subset is a subobject classifier in a topos.  To help build some intuition, consider how to define subsets without ``looking inside" a set. Essentially, a subset $S$ of some larger set $T$ can be viewed as a ``monic arrow", that is, an injective (or 1-1) function $f: S \hookrightarrow T$. Our approach builds on this abstraction to define a category ${\cal C}_{TCM}$ whose objects are causal models, such as SCMs or Markov categories, and a submodel $M_x$ of an SCM $M$ is simply a monic arrow $f_x: M_x \hookrightarrow M$. 

\begin{definition}
    An {\bf elementary topos} is a category ${\cal C}$ that has all (i) limits and (ii) colimits, (iii) has exponential objects, and (iv) a subobject classifier.
\end{definition}
For example, the category of sets forms a topos. Limits exist because one can define Cartesian products of sets, and colimits correspond to forming set unions. Exponential objects correspond to the set of all functions between two sets. Finally, the subobject classifier is simply the subset function, which induces a boolean-valued characteristic function.  A few of the many alternative definitions are to combine (i) and (iii) by the condition that ${\cal C}$ is {\em Cartesian closed}; another reformulation is to replace condition (i) by the condition that ${\cal C}$ has a terminal object and pullbacks; and similarly, condition (ii) can be replaced by saying ${\cal C}$ has an initial object and coequalizers of all maps between coproducts. These are all different ways of characterizing a topos in terms of its universal properties.  Let us begin with the most interesting condition (iv), as it relates closely to the idea of constructing submodels of SCMs by an intervention. Let us remind ourselves of the definitions of subobject classifiers in a topos. See Section~\ref{introcat} in the Appendix for more explanation,  and Table~\ref{setvscategories} for a comparison of commonly defined constructs on sets vs. a topos. 
\begin{definition}
    In a category ${\cal C}$ with finite limits, a {\bf subobject classifier} is a ${\cal C}$-object $\Omega$, and a ${\cal C}$-arrow ${\bf true}: {\bf 1} \rightarrow \Omega$, such that to every other monic arrow $S \hookrightarrow X$ in ${\cal C}$, there is a unique arrow $\phi$ that forms the following pullback square: 
\[\begin{tikzcd}
	S &&& {{\bf 1}} \\
	\\
	X &&& \Omega
	\arrow["m", tail, from=1-1, to=3-1]
	\arrow[from=1-1, to=1-4]
	\arrow["{{\bf true}}"{description}, tail, from=1-4, to=3-4]
	\arrow["{\phi}"{description}, dashed, from=3-1, to=3-4]
\end{tikzcd}\]  
\end{definition}
This commutative diagram is a classic example of how to state a universal property: it enforces a condition that every monic arrow $m$ (i.e., every $1-1$ function) that maps a ``sub"-object $S$ to an object $X$ must be characterizable in terms of a ``pullback", a particular type of universal property that is a special type of a limit. In the special case of the category of sets, it is relatively easy to show that subobject classifiers are simply the characteristic (Boolean-valued) function $\phi$ that defines subsets. In general, as we show below, the subobject classifier $\Omega$ for causal models is not Boolean-valued, and requires using intuitionistic logic through a Heyting algebra.  This definition can be rephrased as saying that the subobject functor is representable. In other words, a subobject of an object $x$ in a category ${\cal C}$ is an equivalence class of monic arrows $m: S \hookrightarrow  x$.

 The fundamental claim made in this paper is that a TCM has the ``right" universal properties to define a general theory of causality. To illustrate this claim, we show in this section that TCMs (and indeed) SCMs do indeed define a topos.  

 \begin{theorem}
 \label{gdcthm}
     The category ${\cal C}_{TCM}$  forms a topos. 
 \end{theorem}
 {\bf Proof Sketch:} The proof essentially involves checking each of the conditions in the above definition of a topos. We will focus on the subobject classifier construction.   The key idea is to observe that any TCM $M = \langle U, V, F \rangle$ induces a function $F: U \rightarrow V$.  It is well-known that such function categories do indeed form toposes.  We give the details below, first beginning to show how categories of functions form a topos, and then apply it to TCMs.  $\qed$

The first task is to define more carefully what the ``arrows" of the category ${\cal C}_{TCM}$ are. The objects are clear: they are simply ``global" functions specifying some unique function $F: U \rightarrow V$. The arrows now are given as commutative diagrams as illustrated below. 
 \begin{center}
 \label{gdcarrow}
\begin{tikzcd}
	U && {U'} \\
	\\
	V && {V'}
	\arrow["h"', from=1-1, to=1-3]
	\arrow["f", from=1-1, to=3-1]
	\arrow["{f'}"', from=1-3, to=3-3]
	\arrow["g", from=3-1, to=3-3]
\end{tikzcd}
 \end{center}
 A commutative diagram, as the term suggests, is a structure showing the equivalence of two paths. Here, the diagram asserts that $g \circ f = f' \circ h$. In the context of our category ${\cal C}_{TCM}$, the arrow $f: U \rightarrow V$ is simply an SCM  $M$, and $f$ is its induced mapping from exogenous to endogenous variables. Similarly, $f'$ is also the induced function mapping exogenous to endogenous variables for another SCM $M'$. The morphisms $h$ and $g$ are functions on SCMs, which transform one causal model into another. In the specific case we are interested in, these functions define causal interventions, but in general, they may be arbitrary functions. Definition~\ref{intervention-scm} in Section~\ref{doc} in the Appendix defining an intervention in an SCM in fact explicitly defines one such (monic) arrow between the model $M$ and its submodel $M_x$. 

\begin{figure}
    \centering
    \hfill 
    \includegraphics[width=0.3\linewidth]{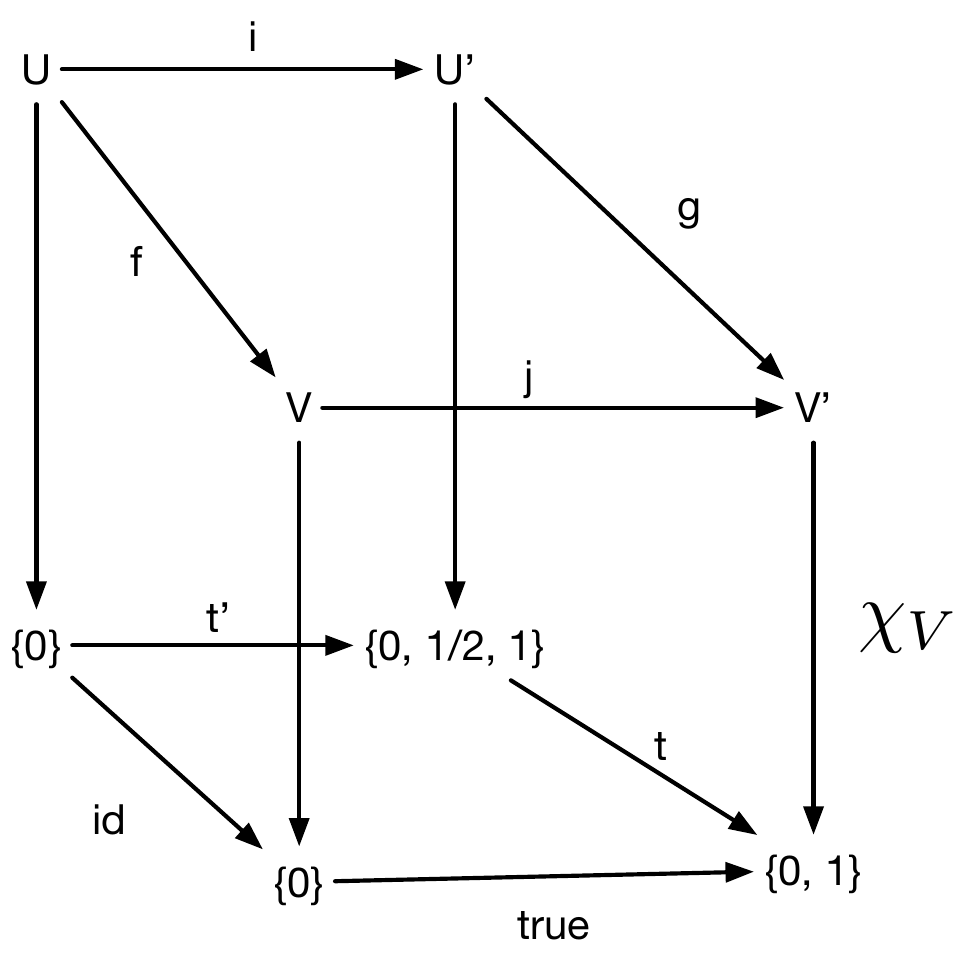}
    \hspace{0.1in}
     \includegraphics[width=0.35\linewidth]{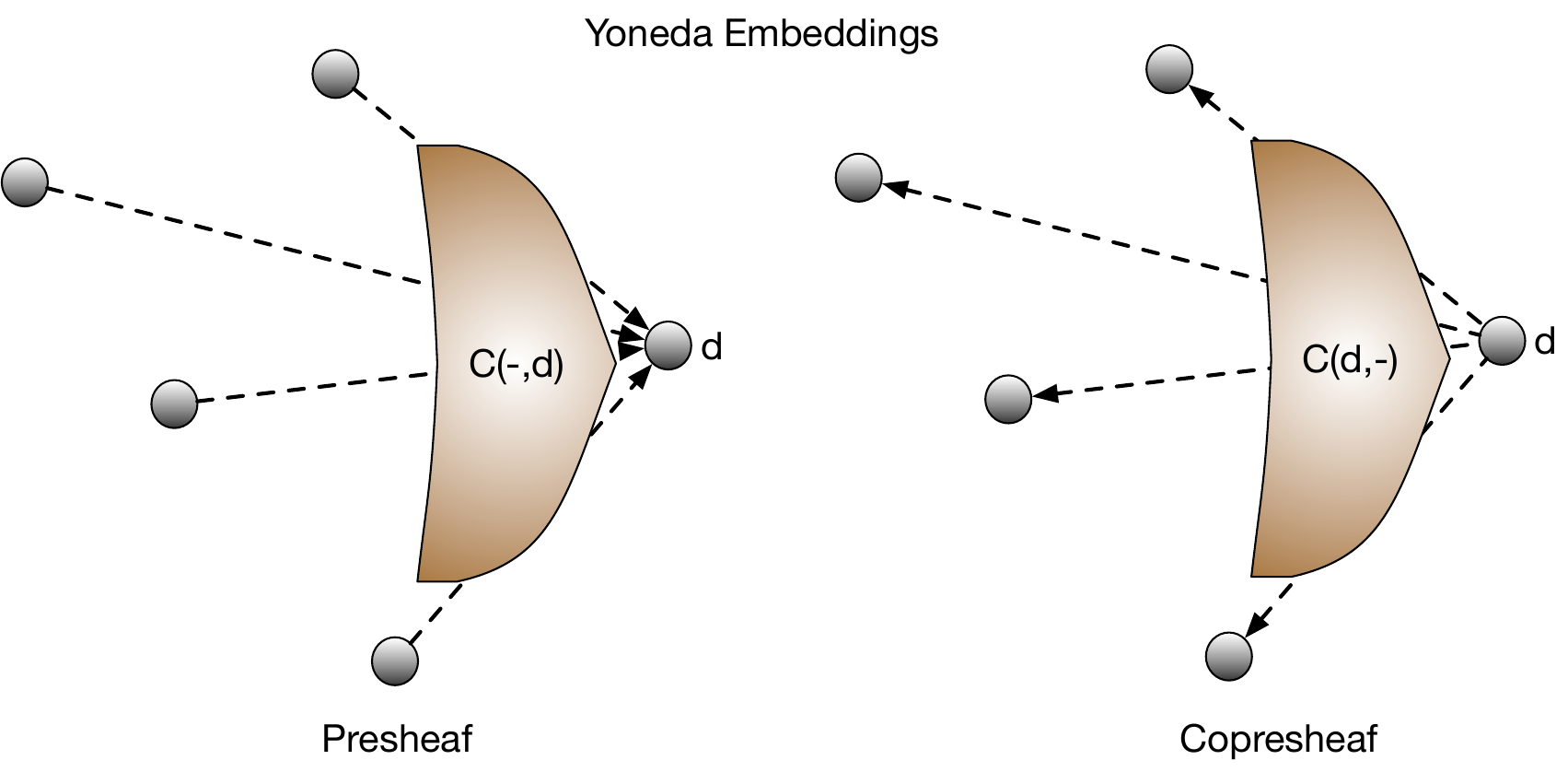}
     \includegraphics[width=0.3\linewidth]{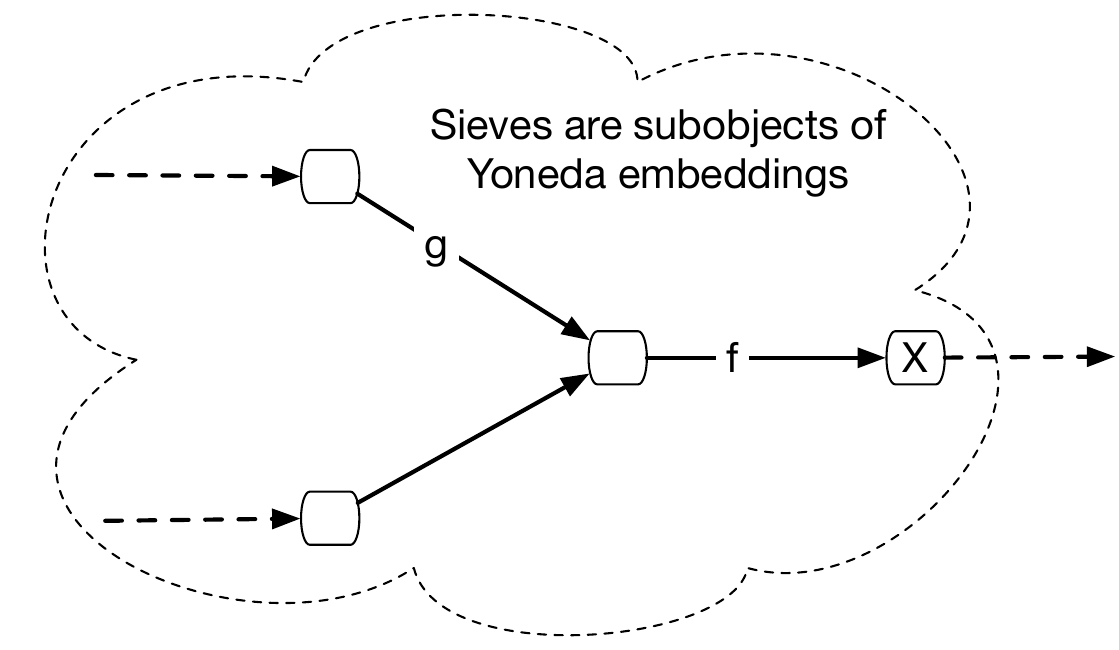}
    \caption{Left: The subobject classifier $\Omega$ for the topos category ${\cal C}_{TCM}$ is displayed on the bottom face of this cube; Right: A TCM can also be constructed from $\yo$ Yoneda embeddings, which maps each object to a set-valued contravariant functor ${\bf Set}^{{\cal C}^{op}}$ called a {\em presheaf}, and then imposing a Grothendieck topology on its sieves -- subobjects of $\yo(x)$ -- that captures underlying causal structure. }
    \label{subobj-classifier}
\end{figure}

Now that we have defined the objects and arrows of the ${\cal C}_{TCM}$ category, let us define the notion of subobject classifiers.  First, we need to define what a ``subobject" is in the category ${\cal C}_{TCM}$. Note that TCMs can abstractly be viewed as functions   $f: U \rightarrow V$,  and $g: U' \rightarrow V'$ etc. Here, let us assume that the TCM $M$ that defines $f$ is a {\em submodel} of the TCM $N$ that induces $g$. We can denote that by defining a commutative diagram as shown below. Let us stress the difference between the diagram shown above for arbitrary functions $g$ and $h$ vs. the one below, where $i$ and $j$ are monic arrows. 
\begin{center}
\label{gdcsubobj}
\begin{tikzcd}
	U && {U'} \\
	\\
	V && {V'}
	\arrow["i", tail, from=1-1, to=1-3]
	\arrow["f"', from=1-1, to=3-1]
	\arrow["g", from=1-3, to=3-3]
	\arrow["j"', tail, from=3-1, to=3-3]
\end{tikzcd}
\end{center} 
Let us examine Figure~\ref{subobj-classifier} to understand the design of subobject classifiers for the category ${\cal C}_{TCM}$. An element $x \in U'$, which is a particular realization of the exogenous variables in $U'$,  can be classified in three ways by defining a characteristic function $\psi$:
\begin{enumerate}
    \item $x \in U$ -- here we set $\psi(x) = {\bf 1}$. 
    \item $x \notin U$ but $g(x) \in V$ -- here we set $\psi(x) = {\bf \frac{1}{2}}$.
    \item $x \notin U$ and $g(x) \notin V$ -- we denote this by $\psi(x) = {\bf 0}$. 
\end{enumerate}

The subobject classifier is illustrated as the bottom face of the cube shown in Figure~\ref{subobj-classifier}: 
\begin{itemize}
    \item ${\bf true}(0) = t'(0) = {\bf 1}$ 
    \item ${\bf t}: {\bf \{0, \frac{1}{2}, 1 \} \rightarrow \{0, 1 \}}$, where ${\bf t(0) = 0}, {\bf t(1) = t(\frac{1}{2}) = 1}$. 
    \item $\chi_V$ is the characteristic function of the exogenous variable set $V$. 
    \item The base of the cube in Figure~\ref{subobj-classifier} displays the subobject classifier ${\bf T: 1 \rightarrow \Omega}$, where ${\bf T} = \langle {\bf t', true} \rangle$ that maps ${\bf 1 = id_{\{0\}}}$ to $\Omega = {\bf t}: \{0, \frac{1}{2}, 1 \} \rightarrow \{0, 1\}$.
\end{itemize}

This proves that the category ${\cal C}_{TCM}$ does not have Boolean semantics, but intuitionistic semantics as its subobject classifier $\Omega$ has multiple degrees of ``truth", corresponding to the three types of classifications of monic arrows (in regular set theory, there are only two classifications). We will relegate the additional details of the proof to Section~\ref{proofs} in the Appendix. \qed 

\section{Topos Causal Models Over  Sheaves}
\label{gdc-stoch}
We now describe a  more general categorical framework for defining TCMs by using the property that Yoneda embeddings of presheaves forms a topos \citep{maclane:sheaves}. Bringing in the theory of sheaves over a topos has the added benefit that it provides an elegant formalism for how to combine independent causal mechanisms, such as individual local functions $f_i$ in SCMs posited in Definition~\ref{scm}, into a consistent unique global function.  We postpone some technical details to the Appendix in Section~\ref{mccat}, and try to give some intuition underlying the categorical machinery. The approach described here generalizes the proof of Theorem~\ref{gdcthm}.  To ensure consistent extension into a unique global function, we build on the theory of sheaves \citep{maclane1992sheaves}, which ensures local functions can be ``collated" together to yield a unique global function. In our setting, we will construct sheaves from categories over causal models through the Yoneda embedding $\yo(x): {\cal C} \rightarrow {\bf Sets}^{{C}^{op}}$ and impose a Grothendieck topology (see Figure~\ref{subobj-classifier}). 

\subsection{TCMs over Graphs}
\label{graph-topos}
\begin{figure}[t]
    \centering
    \includegraphics[width=0.55\linewidth]{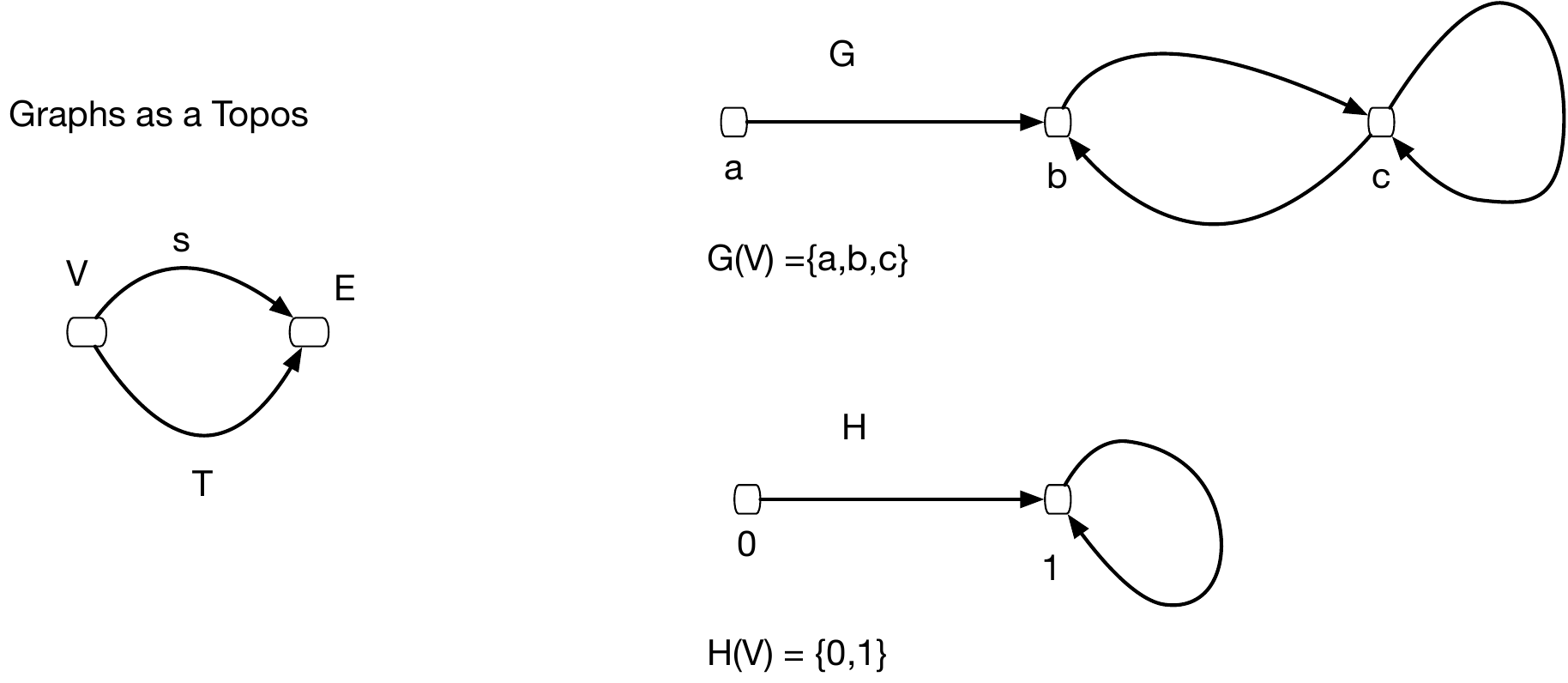}
      \includegraphics[width=0.35\linewidth]{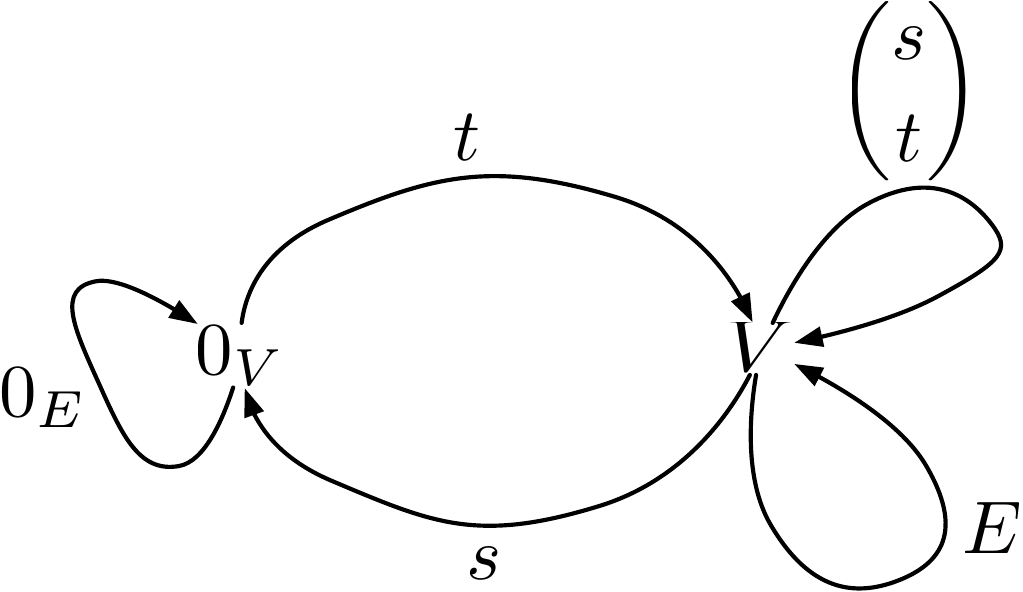}
    \caption{TCMs can be defined over arbitrary directed graphs define a topos. The subobject classifier is illustrated on the right.}
    \label{fgraphtopos}
\end{figure}
To begin with a relatively elementary construction involving sheaves, Figure~\ref{fgraphtopos} illustrates how directed graphs, widely used in causal modeling, can be modeled as a topos that only has two objects: a generic ``vertex" object $V$, representing an abstract causal variable, and a generic ``edge" object $E$, denoting an abstract causal path. Any actual graph, such as the two shown in the figure, are covariant functors from ${\cal C}_\Gamma$, the graph topos, to the actual graphs (we can treat graphs equivalently as contravariant functors by reversing the edges from $V$ to $E$). Sample object mappings of $V$ for the two graphs are shown.  For the topos category ${\cal C}_\Gamma$, the ``representable functor" is defined as the presheaf ${\cal C}_\Gamma(-, c)$ for each object $c$ in the category $\Gamma$, which means the set of all morphisms going {\em into} object $c$.  Let us calculate the representable functors for the topos of graphs ${\cal C}_\Gamma$. Since ${\cal C}_\Gamma$ has only two objects, $V$ and $E$, the representable functors are given as the sets: 
\[ {\cal C}_\Gamma(-, V), \ \ \ \ {\cal C}_\Gamma(-, E) \]
Since ${\cal C}_\Gamma$ has no arrows from object $E$ to object $V$, we can easily check that the representable functor ${\cal C}_\Gamma(-,V)$ is given as: 
\[ {\cal C}_\Gamma(V, V) = \{1_V \},  \  \ \ {\cal C}_\Gamma(E, V) = \emptyset \]
where $1_V$ is the self-loop arrow that maps object $V$ to itself. On the other hand, the representable functor ${\cal C}_\Gamma(-, E)$ is defined as: 
\[ {\cal C}_\Gamma(V, E) = \{s, t \},  \ \ \ \ {\cal C}_\Gamma(E, E) = \{1_E \} \]
Additional details, like constructing exponential objects $H^G$ for the ${\cal C}_\Gamma$ category, are given in \citep{vigna2003guidedtourtoposgraphs}. For the subobject classifier, the idea is as follows. For any causal model $N$ represented as a submodel of a more complex causal model $M$, defined by a monic arrow $m: N \hookrightarrow M$, the generalization of the usual set-theoretic characteristic function is the classifying map $\chi_{m}: N \rightarrow \Omega$. As in Theorem~\ref{gdcthm}, $\Omega$ is not Boolean, but has multiple ``degrees of truth" (see Figure~\ref{fgraphtopos}): 
\begin{enumerate}
    \item Causal variables not in $N$ are mapped to $O_V$. 
    \item Causal variables in $N$ are mapped to $V$. 
    \item If an edge is not in $N$, four cases emerge: 
    \begin{enumerate}
        \item An arc whose source and target are not in $N$ is mapped to $0_E$. 
        \item An arc whose source is in $N$, but target is not, is mapped to an edge $s$. 
        \item An arc whose target is in $N$, but source is not, is mapped to an edge $t$. 
        \item An arc having both source and target in $N$ is mapped to $s \choose t$.\footnote{${s \choose t}: V + V \rightarrow E$ is a map defined by the universal property of coproducts. See Supplementary Materials.}
    \end{enumerate}
\end{enumerate}
\subsection{Grothendieck Topology on Sites}
We introduce a two-step ``toposification" process to construct TCMs: (i) construct the $\yo$ Yoneda embedding of ``presheaves" of some category of causal models,   which maps each object $x$ into a contravariant set-valued functor ${\cal C}(-, x)$. (ii) Use the Grothendieck topology to obtain a unique causal model over sheaves. Figure~\ref{fgraphtopos} gives the high-level intuition, which we can now apply to causal categorical models,  such as an affine CDU ("copy-delete-uniform") category \citep{string-diagram-surgery}, or a Markov category \citep{fritz:jmlr}, or a simplicial set \citep{DBLP:journals/entropy/Mahadevan23}, as well as Lewis' theory of counterfactuals \citep{DBLP:journals/jphil/Lewis73}.  

%
\begin{definition}\citep{maclane1992sheaves}
\label{sheaftopology}
    A {\bf sheaf} of sets $F$ on a topological space $X$ is a functor $F: {\cal O}^{op} \rightarrow {\bf Sets} $ such that each open covering $U = \bigcup_i U_i, i \in I$ of an open set $O$ of $X$ yields an equalizer diagram
    \[ FU \xrightarrow[]{e} \prod_i FU_i \doublerightarrow{p}{q} \prod_{i,j} F(U_i \cap U_j) \]
\end{definition}
To concretize this categorical machinery, consider the special case where $FU = \{ f | f: U \rightarrow \mathbb{R} \}$ is the set of all continuous real-valued functions. Then the sheaf condition in Definition~\ref{sheaftopology} states that any restriction of $F$ on overlapping subspaces $FU_i$ and $FU_j$ must be consistent over their intersection $U_i \cap U_j$ by asserting the existence of an ``equalizer" arrow $e$. Exploiting the property that an elementary topos always has equalizers, we can construct sieves in causal categories over a topos, but we need to generalize the definition above. 
\begin{definition}
A {\bf sieve} for any object $x$ in any (small) category ${\cal C}$ is a subobject of its Yoneda embedding $\yo(x) = {\cal C}(-,x)$. If $S$ is a sieve on $x$, and $h: D \rightarrow x$ is any arrow in category ${\cal C}$, then 
    \[ h^*(S) = \{g \ | \ \mbox{cod}(g) = D, hg \in S \}\]
\end{definition}
\begin{definition}\citep{maclane1992sheaves}
A {\bf Grothendieck topology} on a category ${\cal C}$ is a function $J$ which assigns to each object $x$  of ${\cal C}$ a collection $J(x)$ of sieves on $x$ such that
\begin{enumerate}
    \item the maximum sieve $t_x = \{ f | \mbox{cod}(f) = x \}$ is in $J(x) $. 
    \item If $S \in J(x)$  then $h^*(S) \in J(D)$ for any arrow $h: D \rightarrow x$. 
    \item If $S \in J(x)$ and $R$ is any sieve on $x$, such that $h^*(R) \in J(D)$ for all $h: D \rightarrow x$, then $R \in J(C)$. 
\end{enumerate}
\end{definition}
We can now define categories with a given Grothendieck topology as {\em sites}. 
\begin{definition}
    A {\bf site} is defined as a pair $({\cal C}, J)$ consisting of a small category ${\cal C}$ and a Grothendieck topology $J$ on ${\cal C}$. 
\end{definition}
\begin{definition}
\label{subobj-defn}
    The {\bf subobject classifier} $\Omega$ is defined on any topos ${\bf Sets}^{C^{op}}$ as subobjects of the representable functors: 
    \[ \Omega(X) = \{S | S \ \ \mbox{is a subobject of} \ \ {\cal C}(-, X) \} \]
    and the morphism ${\bf true}: 1 \rightarrow \Omega$ is ${\bf true}(X) = X$ for any representable $X$ 
\end{definition}

\subsection{TCMs over Functor Categories}

We discuss how to define TCMs over categories of functors. For example, the category of all Bayesian network can be modeled as a functor category, as shown in \citep{string-diagram-surgery,fritz:jmlr}. 
\begin{theorem}
The  functor category over Bayesian Networks defined as CDU functors from the affine ``copy-delete-uniform" (CDU) category to the symmetric monoidal category {\bf FinStoch} of finite stochastic processes forms a TCM. 
\end{theorem}
{\bf Proof Sketch:} \citet{string-diagram-surgery} showed that Bayesian Networks can be modeled as functors from the symmetric monoidal category called CDU (for ``copy-delete-uniform") to the symmetric monoidal category {\bf FinStoch} of finite stochastic processes. Causal interventions on Bayesian Networks in terms of ``graph surgery" can be exactly reproduced as ``string diagram surgery".  The functor category whose  objects are CDU functors forms a topos. The proof is simply a special case of the general proof that the Yoneda embedding of presheafs ${\cal C}^{op}(-,x)$ forms a topos \citep{maclane1992sheaves}. \qed 

\begin{theorem}
The  category of causal models defined as presheaves over Markov categories forms a TCM.
\end{theorem}

{\bf Proof Sketch}: \citep{Fritz_2020} showed that Markov categories, which are symmetric monoidal categories with a comonoidal ``copy-delete" structure on each object, provides a unifying foundation for probability  and statistics. Arrows are causal interventions as ~`surgery" on string diagrams, as shown in \citep{string-diagram-surgery}. By constructing the Yoneda embeddings of Markov categories, we can construct a topos over Markov categories, where the arrows correspond to string diagram surgery, as defined in \citep{string-diagram-surgery}. \qed 

\citet{DBLP:journals/entropy/Mahadevan23} studied causal inference over {\em simplicial sets} \citep{may1992simplicial}, which are combinatorial structures in algebraic topology that define CW-complexes (compact generated weak-Hausdorff spaces). 
\begin{theorem}
    The category over causal models represented as simplicial sets forms a TCM. 
\end{theorem}
{\bf Proof:}  A simplicial set is a graded set $X_n, n \geq 0$, where $X_0$ can be defined as a collection of ``objects" (e.g., SCMs), $X_1$ defines $1$-simplices that can be viewed as arrows $M \rightarrow N$ of morphisms between causal models $M$ and $N$, and so on. Simplicial sets have a natural topological realization as a CW-complex \citep{milnor}. It is a standard result in abstract homotopy theory that simplicial sets form a topos \citep{Lurie:higher-topos-theory}. \qed 


\section{Internal Logic of a TCM}

Finally, we introduce the third major component of our TCM framework: an internal formal language called the Mitchell-B\'enabou language (MBL) that enables logical causal and counterfactual reasoning. We introduce Heyting algebras associated with the structure of subobject classifiers (which in general are defined according to Definition~\ref{subobj-defn}).  We define the MBL associated with a causal topos  and specify its Kripke-Joyal semantics.  

\subsection{Heyting Algebras} 

\begin{definition}
    A {\bf Heyting algebra} is a poset with all finite products and coproducts, which is Cartesian closed. That is, a Heyting algebra is a lattice, including bottom and top elements, denoted by ${\bf 0}$ and ${\bf 1}$, respectively, which associates to each pair of elements $x$ and $y$ an exponential $y^x$. The exponential is written $x \Rightarrow y$, and defined as an adjoint functor: 
    \[ z \leq (x \Rightarrow y) \ \ \mbox{if and only if} \ \ z \wedge x \leq y\]
\end{definition}
In other words, $x \Rightarrow y$ is a least upper bound for all those elements $z$ with $z \wedge x \leq y$.  As a concrete example, for a topological space $X$ the set of open sets ${\cal O}(X)$ is a Heyting algebra. The binary intersections and unions of open sets yield open sets. The empty set $\emptyset$ represents ${\bf 0}$ and the complete set $X$ represents ${\bf 1}$. Given any two open sets $U$ and $V$, the exponential object $U \Rightarrow W$ is defined as the union $\bigcup_i W_i$ of all open sets $W_i$ for which $W \cap U \subset V$. 


Note that in a Boolean algebra, we define implication as the relationship $(x \Rightarrow y) \equiv \neg x \vee y$. This property is referred to as the ``law of the excluded middle" (because if $x = y$, then this translates to $\neg x \vee x = {\bf true}$) does not always hold in a Heyting algebra. 
%

\subsection{Mitchell-B\'enabou Language} 

We briefly sketch the construction of the internal logic of a topos here.   The  Mitchell-B\'enabou language (MBL) is a typed local set theory (see Section~\ref{lst}) associated with a causal topos. Given the topos category ${\cal C}_{TCM}$, we define the types of MBL as causal model objects $M$ of ${\cal C}_{TCM}$. For each type $M$ (e.g., a TCM),  we assume the existence of variables $x_M, y_M, \ldots$, where each such variable has as its interpretation the identity arrow ${\bf 1}: M \rightarrow M$. We can construct product objects, such as $A \times B \times C$, where terms like $\sigma$ that define arrows are given the interpretation $\sigma: A \times B \times C \rightarrow D$. We can inductively define the terms and their interpretations in a topos category as follows (see \citep{maclane1992sheaves} for additional details). Here are two cases, the remainder are detailed in Section~\ref{mbl} in the Appendix. 

\begin{itemize}
    \item Each variable $x_M$ of type $M$ is a term of type $M$, and its interpretation is the identity $x_M = {\bf 1}: M \rightarrow M$ (e.g., $M$ may be a TCM over a  Markov category). 

\item Terms $\sigma$ and $\tau$ of types $C$ and $D$ that are interpreted as $\sigma: A \rightarrow C$ and $\tau: B \rightarrow D$ can be combined to yield a term $\langle \sigma, \tau \rangle$ of type $C \times D$, whose joint interpretation is given as 

\[ \langle \sigma p, \tau q \rangle: X \rightarrow C \times D\]

where $X$ has the required projections $p: X \rightarrow A$  and $q: X \rightarrow B$. 

\end{itemize} 

\subsection{Kripke-Joyal Semantics for a Causal Topos}

We briefly sketch the semantics for the Mitchell-B\'enabou language here using the Kripke-Joyal possible worlds construction \citep{maclane:sheaves}, and give the full details in Section~\ref{kj} in the Appendix. Any free variable $x$ must have some causal model $X$ of ${\cal C}_{TCM}$ as its type.  For any causal model $M$ in ${\cal C}_{TCM}$, define a {\em generalized element} as a morphism $\alpha: N \rightarrow M$. To understand this definition, note that we can define an element of a causal model by the  morphism $x: {\bf 1} \rightarrow M$. Thus, a generalized element $\alpha: N \rightarrow M$ represents the ``stage of definition" of $M$ by $N$. We specify the semantics of how an TCM $N$ supports any formula $\phi(\alpha)$, denoted by $N \Vdash \phi(\alpha)$, as follows: 
\[ N \Vdash \phi(\alpha) \ \ \ \mbox{if and only if } \ \ \ \mbox{Im} \ \alpha \leq \{ x | \phi(x) \} \]
Stated in the form of a commutative diagram, this ``forcing" relationship holds if and only if $\alpha$ factors through $\{x | \phi(x) \}$, where $x$ is a variable of type $M$ (recall that objects $M$ of a topos form its types), as shown in the following commutative diagram. 
 \begin{center}
\begin{tikzcd}
	&& {\{x | \phi(x) \}} && {{\bf 1}} \\
	\\
	N && M && {{\bf \Omega} }
	\arrow[from=1-3, to=1-5]
	\arrow[tail, from=1-3, to=3-3]
	\arrow["{{\bf True}}", from=1-5, to=3-5]
	\arrow[dashed, from=3-1, to=1-3]
	\arrow["\alpha"', from=3-1, to=3-3]
	\arrow["{\phi(x)}", from=3-3, to=3-5]
\end{tikzcd}
 \end{center}

\section{Summary of Our Framework}
\label{limitations}

In this paper, we proposed topos causal models (TCMs), a novel class of causal models that exploit the key properties of a topos category: they are (co)complete, meaning  all (co)limits exist, they admit a {\em subobject classifier}, and allow {\em exponential objects}. The main goal of this paper is to show that these properties are central to many applications in causal inference. For example, subobject classifiers allow a categorical formulation of causal intervention, which creates sub-models. Limits and colimits allow causal diagrams of arbitrary complexity to be ``solved", using a novel interpretation of causal approximation. Exponential objects enable reasoning about equivalence classes of operations on causal models, such as covered edge reversal and causal homotopy. Analogous to structural causal models (SCMs), TCMs are defined by a collection of functions, each defining a ``local autonomous" causal mechanism that assemble to induce a unique global function from exogenous to endogenous variables. Since the category of TCMs is (co)complete, which we prove in this paper, every causal diagram has a ``solution" in the form of a (co)limit: this implies that any arbitrary causal  model can be ``approximated" by some global function with respect to the morphisms going into or out of the diagram. Natural transformations are crucial in measuring the quality of approximation. In addition, we show that causal interventions are modeled by subobject classifiers: any sub-model is defined by a monic arrow into its parent model. Exponential objects permit reasoning about entire classes of causal equivalences and interventions. Finally, as TCMs form a topos, they admit an internal logic defined as a Mitchell-Benabou language with an associated Kripke-Joyal semantics. We show how to reason about causal models in TCMs using this internal logic.  

There are many directions for future work, some of which we discuss below. 

\begin{itemize}
    \item {\em Categorical homotopy in TCM}: It is well-known that causal models can only be identified up to an equivalence class from observational data, without additional (perhaps impossible) interventions \citep{pearl:bnets-book,spirtes:book}. We can use the framework of categorical homotopy theory to reason about equivalence classes of TCM objects \citep{Quillen:1967}. In our recent work \citep{cktheory}, we proposed the use of Quillen's higher algebraic K-theory as a means to reason about causal equivalence. This framework can also be applied to TCMs. 

    \item {\em Internal logic of a TCM:} We sketched out the internal logic of a TCM, but we did not use it to prove interesting properties of TCM models. One example of using internal logic of a topos to prove properties of Lewis-style counterfactuals \citep{DBLP:journals/jphil/Lewis73} is given in \citep{topos-counterfactual}. This topos-theoretic framework of counterfactuals can be compared wtih Pearl's axiomatic theory as well \citep{scm-lewis}. 

    \item {\em Approximating TCMs by (co)limits}: One intriguing aspect of our TCM framework is the use of (co)limits to solve functor diagrams specifying causal models. Using the completeness of the TCM category implies that any causal diagram is ``solvable", meaning that it can be approximated by a TCM object with respect to all morphisms going into the diagram (limits) or going out of the diagram (colimits). Investigating the properties of universal constructions, such as limits, colimits, and more generally, Kan extensions is an interesting avenue for future work. 
    
\end{itemize}


\newpage 

\section{Appendix: Causal Inference Background}

We begin in Section~\ref{doc} with a brief summary of Pearl's structural causal models (SCMs), which induces a topos as shown in the main paper. We also briefly discuss the application of TCMs to counterfactual inference in Section~\ref{lewis}. We give an introduction to categories and functors in Section~\ref{introcat}. Section~\ref{toposes} gives a brief overview of the theory of sheaves over toposes. Section~\ref{lst} defines local set theories, another way to characterize the internal language of a topos.  Section~\ref{mccat} contains an overview of affine CDU and Markov categories, which have been previously studied as categorical models of causality and probability \citep{string-diagram-surgery,fritz:jmlr}. Finally, in Section~\ref{proofs}, we include more detailed proofs of the main theorems in the paper.

\subsection{Do-Calculus}
\label{doc} 

We review the notion of a structural causal model (SCM) \citep{pearl-book}. Succinctly, any SCM $M$ defines a unique function from exogenous variables to endogenous variables, and do-calculus models interventions as ``sub-functions": 

\begin{definition}\citep{pearl-book}
\label{scm}
    A {\bf structural causal model} (SCM) is defined as the triple $\langle U, V, F \rangle$ where $V = \{V_1, \ldots, V_n \}$ is  a set of {\em endogenous} variables, $U$ is a set of {\em exogenous} variables, $F$ is a set $\{f_1, \ldots, f_n \}$ of ``local functions" $f_i: U \cup (V \setminus V_i) \rightarrow V_i$ whose composition induces  a unique function $F$ from  $U$ to $V$. 
\end{definition}
\begin{definition}\citep{pearl-book}
\label{intervention-scm}
   Let $M = \langle U, V, F \rangle $ be a causal model defined as an SCM, and $X$ be a subset of variables in $V$, and $x$ be a particular realization of $X$.  A {\bf submodel} $M_x = \langle U, V, F_x \rangle $ of $M$ is the causal model $ M_x =  \langle U, V, F_x \rangle$, where $F_x = \{f_i : V_i \notin X \} \cup \{X = x \}$. 
\end{definition}
\begin{definition}\citep{pearl-book}
    \label{do-action}
    Let $M$ be an SCM, $X$ be a set of variables in $V$, and $x$ be a particular realization of $X$. The {\bf effect} of an action $\mbox{do}(X=x)$ on $M$ is given by the submodel $M_x$. 
\end{definition}
\begin{definition}\citep{pearl-book}
    \label{potential-outcome}
    Let $Y$ be a variable in $V$, and let $X$ be a subset of $V$. The {\bf potential outcome} of $Y$ in response to an action $\mbox{do}(X=x)$, denoted $Y_x(u)$, is the solution of $Y$ for the set of equations $F_x$. 
\end{definition}
\citet{scm-lewis} propose an axiomatic theory of counterfactuals based on the above definitions, where the key definition of a counterfactual is given as:
\begin{definition}
    Let $Y$ be a variable in $V$ and let $X$ be a subset of $V$. The counterfactual sentence ``The value that $Y$ would have obtained had $X$ been set to $x$" is defined as the potential outcome $Y_x(u)$.
\end{definition}

\subsection{Counterfactuals using TCMs}
\label{lewis}
Finally, let us briefly indicate how to construct counterfactuals in a topos category defined over causal models, which generalizes the notion of potential outcome in Definition~\ref{potential-outcome}.  \citet{topos-counterfactual} describe an intuitionistic logic for Lewis' \citep{DBLP:journals/jphil/Lewis73} theory of counterfactuals, where the neighborhood system of possible worlds is governed by the graph topos ${\cal C}_\Gamma$ illustrated in Figure~\ref{fgraphtopos}. However,  as   \citet{scm-lewis}  argued,  many  counterfactuals have causal meaning. For example, the well-known counterfactual ``If Kangaroos had no tails, they would fall over" attains causal meaning as the logic of counterfactuals proposed by Lewis evaluates its truth in the nearest possible worlds, where the laws of physics are the same as our world, except for  a ``kangaroo surgery".  Lewis introduces two counterfactual connectives, and we illustrate how to model one of them in our framework:
\begin{itemize}
    \item $\alpha  \boxright \beta$ is true in a world $u$ according to a neighborhood system ${\cal C}_{TCM}$, if either for no world $w$ in ${\cal C}_{TCM}(u)$, $\models_w \alpha$, or there is some neighborhood $N$ in ${\cal C}_{TCM}(u)$  that has a world $w$ such that $\models_w \alpha$ and $\models_v \alpha \Rightarrow \beta$ in every world $v$ of $N$. 
\end{itemize}
Crucially, unlike the case for \citep{topos-counterfactual}, our topos-theoretic framework imposes a causal structure on the neighborhood system. We next describe how to evaluate causal or counterfactual logical statements of the above form in an internal intuitionistic logic of a causal topos.

\section{Category Theory Background}

\label{introcat}

\begin{figure}[!htb]
  \centering
  \includegraphics[width=3.5in]{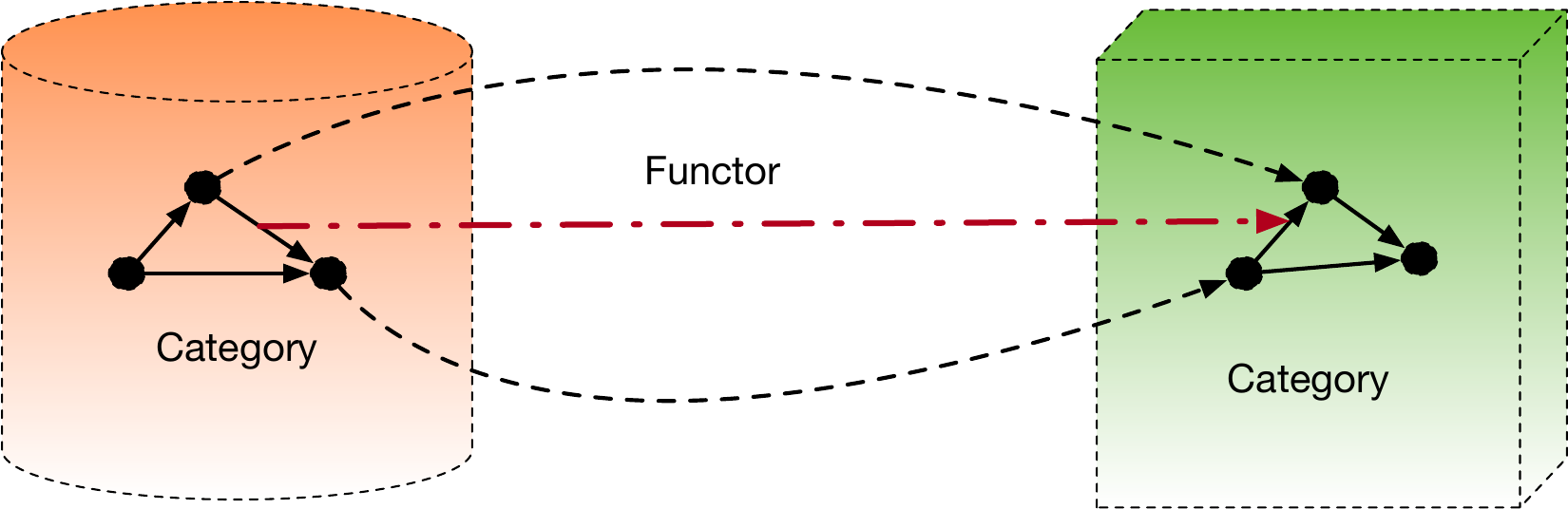}
  \caption{Categories are collections of objects, with a collection of arrows defined between each pair. A functor must map both objects and arrows from a domain category into a co-domain category.}\label{fig:cat}
\end{figure}

Category theory is perhaps the most transformative rethinking of mathematics since antiquity. Rather than focus on the internals of an object, like the elements of a set, it focuses on the arrows or morphisms that define the interactions between objects. Applied to causal models, it suggests that we can construct languages where individual causal models can be defined as variables in a logical language with intuitionistic semantics. The Yoneda Lemma, one of the most celebrated results in pure mathematics of the 20th century, states that objects are completely defined up to isomorphism purely in terms of their interactions. Applied to SCMs, its implications are nonetheless startling. SCMs in effect can be defined not just in terms of their internal structure, but in terms of how they  interact with other SCMs in a category of such objects. Being conditioned to think in terms of internal structure, this statement may seem counterintuitive. But, as category theory has shown in numerous cases, such as in algebraic topology \citet{may1992simplicial}, we can often understand deep properties of complex objects such as topological spaces by modeling them as combinatorial objects and analyzing their interactions. 

Our framework builds on the theory of categories, functors, and natural transformations \citep{maclane:71,riehl2017category} (see Figure~\ref{fig:cat} and Figure~\ref{setvscategories}). Many common mathematical structures -- groups, rings, modules, measurable spaces, topological spaces etc. -- form categories. More interesting categories for AI are those associated with compositional machine learning models, such as deep learning \citep{DBLP:conf/lics/FongST19}, where a  symmetric monoidal category {\tt Learn} is defined that combines an {\tt Implement} routine that maps an input $A$ into some output $B$ parameterized by some parameter $P$, and an {\tt Update} routine that given an input-output pair $A,B$ and a parameter $P$, returns a new parameter $P$. Causal models, such as DAGs, structural equation models etc. can be straightforwardly mapped into categories. We will not define the semantics of categories formally, except to say that each object $x$ in a category ${\cal C}$ has an associated identity arrow ${\bf 1}_x$ that maps from $x$ to $x$, and that arrows compose, so that if $f: x \rightarrow y$ and $g: y \rightarrow z$, then $g \circ f: x \rightarrow z$. 

\begin{figure}[t]
\begin{center}
\begin{tabular}{|c |c | } \hline 
{\bf Set theory } & {\bf Topos theory} \\ \hline 
 set & object \\ 
 subset & subobject \\
 truth values $\{0, 1 \} $ & subobject classifier $\Omega$ \\
power set $P(A) = 2^A$ & power object $P(A) = \Omega^A$ \\ \hline
bijection & isomorphims \\ 
injection & monic arrow \\
surjection & epic arrow \\ \hline
singleton set $\{ * \}$ & terminal object ${\bf 1}$ \\ 
empty set $\emptyset$ & initial object ${\bf 0}$ \\
elements of a set $X$ & morphism $f: {\bf 1} \rightarrow X$ \\
- & functors, natural  transformations \\ 
- & limits, colimits, adjunctions \\ \hline
\end{tabular}
\end{center}
\caption{A topos is a category that generalizes set theory: subsets become subobjects and the characteristic function for a subset, which is Boolean, turns into a subobject classifier that may have multiple ``degrees of truth".} 
\label{setvscategories}
\end{figure} 

The celebrated Yoneda Lemma \citep{maclane:71} states that any set-valued functor $F: {\cal C} \rightarrow {\bf Sets}$ can be modeled in terms of natural transformations between $F$ and the Yoneda embedding ${\cal C}(x,-)$ (or ${\cal C}(-,x)$) of a {\em representable} functor. In essence the action of a set valued functor $F$ on an object $x$ is completely determined by its natural transformations with ${\cal C}(-, x)$. What are natural transformations? These specify how functors interact, much like how objects interact through arrows.  A remarkable property of category theory is that the set of all sets is not a set, but the category of all categories is indeed a category, where the arrows are functors, and the objects are categories. 

\begin{definition}
    Given categories $C$ and $D$, and functors $F, G: C \rightarrow D$, a {\bf natural transformation} $\alpha: F \Rightarrow G$ is defined by the following data: 

    \begin{itemize}
        \item an arrow $\alpha_c: Fc \rightarrow Gc$ in $D$ for each object $c \in C$, which together define the components of the natural transformation. 
        \item For each morphism $f: c \rightarrow c'$, the following commutative diagram holds true: 

\[\begin{tikzcd}
	Fc &&& Gc \\
	\\
	{Fc'} &&& {Gc'}
	\arrow["{\alpha_c}", from=1-1, to=1-4]
	\arrow["Ff"', from=1-1, to=3-1]
	\arrow["{\alpha_{c'}}"', from=3-1, to=3-4]
	\arrow["Gf", from=1-4, to=3-4]
\end{tikzcd}\]

    \end{itemize}
    A {\bf natural isomorphism} is a natural transformation $\alpha: F \Rightarrow G$ in which every component $\alpha_c$ is an isomorphism. 
\end{definition}

\subsection{Yoneda Lemma}

The Yoneda Lemma states that the set of all morphisms into an object $d$ in a category ${\cal C}$, denoted as {\bf Hom}$_{\cal C}(-,d)$ and called the {\em contravariant functor} (or presheaf),  is sufficient to define $d$ up to isomorphism. The category of all presheaves forms a {\em category of functors}, and is denoted $\hat{{\cal C}} = $ {\bf Set}$^{{\cal C}^{op}}$. The Yoneda lemma plays a crucial role in this paper because it defines the concept of a {\em universal representation} in category theory. We first show that associated with universal arrows is the corresponding induced isomorphisms between {\bf {Hom}} sets of morphisms in categories. This universal property then leads to the Yoneda lemma. 

\begin{center}
 \begin{tikzcd}
  D(r,r) \arrow{d}{D(r, f')} \arrow{r}{\phi_r}
    & C(c, Sr) \arrow[]{d}{C(c, S f')} \\
  D(r,d)  \arrow[]{r}[]{\phi_d}
&C(c, Sd)\end{tikzcd}
 \end{center} 

 As the two paths shown here must be equal in a commutative diagram, we get the property that a bijection between the {\bf {Hom}} sets holds precisely when $\langle r, u: c \rightarrow Sr \rangle$ is a universal arrow from $c$ to $S$. Note that for the case when the categories $C$ and $D$ are small, meaning their {\bf Hom} collection of arrows forms a set, the induced functor {\bf {Hom}}$_C(c, S - )$ to {\bf Set} is isomorphic to the functor {\bf {Hom}}$_D(r, -)$. This type of isomorphism defines a universal representation, and is at the heart of the causal reproducing property (CRP) defined below. 

\begin{lemma}
{\bf {Yoneda lemma}}: For any functor $F: C \rightarrow {\bf Set}$, whose domain category $C$ is ``locally small" (meaning that the collection of morphisms between each pair of objects forms a set), any object $c$ in $C$, there is a bijection 

\[ \mbox{Hom}(C(c, -), F) \simeq Fc \]

that defines a natural transformation $\alpha: C(c, -) \Rightarrow F$ to the element $\alpha_c(1_c) \in Fc$. This correspondence is natural in both $c$ and $F$. 
\end{lemma}

There is of course a dual form of the Yoneda Lemma in terms of the contravariant functor $C(-, c)$ as well using the natural transformation $C(-, c) \Rightarrow F$. A very useful way to interpret the Yoneda Lemma is through the notion of universal representability through a covariant or contravariant functor.

\begin{definition}
    A {\bf universal representation} of an object $c \in C$ in a category $C$ is defined as a contravariant functor $F$ together with a functorial representation $C(-, c) \simeq F$ or by a covariant functor $F$ together with a representation $C(c, -) \simeq F$. The collection of morphisms $C(-, c)$ into an object $c$ is called the {\bf presheaf}, and from the Yoneda Lemma, forms a universal representation of the object. 
\end{definition}

\subsection{Universal Properties in Categories}

A fundamental principle of category theory is to characterize objects by universal properties. Take the Cartesian product of two sets. The conventional way to define this product of two sets is as the set of ordered pairs, one drawn from each set. But this definition does not specify its universal property. We explain how to define universal properties below, which will be essential to understanding a topos.  A key distinguishing feature of category theory is the use of diagrammatic reasoning. However, diagrams are also viewed more abstractly as functors mapping from some indexing category to the actual category. Diagrams are useful in understanding universal constructions, such as limits and colimits of diagrams. To make this somewhat abstract definition concrete, let us look at some simpler examples of universal properties, including co-products and quotients (which in set theory correspond to disjoint unions). Coproducts refer to the universal property of abstracting a group of elements into a larger one.

 Before we formally the concept of limit and colimits, we consider some examples.  These notions generalize the more familiar notions of Cartesian products and disjoint unions in the category of {\bf {Sets}}, the notion of meets and joins in the category {\bf {Preord}} of preorders, as well as the  least upper bounds and greatest lower bounds in lattices, and many other concrete examples from mathematics. 

\begin{example} 
If  we consider a small  ``discrete'' category ${\cal D}$ whose only morphisms are identity arrows, then the colimit of a functor ${\cal F}: {\cal D} \rightarrow {\cal C}$ is the {\em categorical coproduct} of ${\cal F}(D)$ for $D$, an object of category {\cal D}, is denoted as 
\[ \mbox{Colimit}_{\cal D} F = \bigsqcup_D {\cal F}(D) \]

In the special case when the category {\cal C} is the category {\bf {Sets}}, then the colimit of this functor is simply the disjoint union of all the sets $F(D)$ that are mapped from objects $D \in {\cal D}$. 
\end{example} 

\begin{example} 
Dual to the notion of colimit of a functor is the notion of {\em limit}. Once again, if we consider a small  ``discrete'' category ${\cal D}$ whose only morphisms are identity arrows, then the limit of a functor ${\cal F}: {\cal D} \rightarrow {\cal C}$ is the {\em categorical product} of ${\cal F}(D)$ for $D$, an object of category {\cal D}, is denoted as 
\[ \mbox{limit}_{\cal D} F = \prod_D {\cal F}(D) \]

In the special case when the category {\cal C} is the category {\bf {Sets}}, then the limit of this functor is simply the Cartesian product of all the sets $F(D)$ that are mapped from objects $D \in {\cal D}$. 
\end{example} 

Category theory relies extensively on {\em universal constructions}, which satisfy a universal property. One of the central building blocks is the identification of universal properties through formal diagrams.  Before introducing these definitions in their most abstract form, it greatly helps to see some simple examples.  We can illustrate the limits and colimits in diagrams using pullback and pushforward mappings.
\begin{center}
\begin{tikzcd}
    & Z\arrow[r, "p"] \arrow[d, "q"]
      & X \arrow[d, "f"] \arrow[ddr, bend left, "h"]\\
& Y \arrow[r, "g"] \arrow[drr, bend right, "i"] &X \sqcup Y \arrow[dr, "r"]  \\ 
& & & R 
\end{tikzcd}
\end{center}
An example of a universal construction is given by the above commutative diagram, where the coproduct object $X \sqcup Y$ uniquely factorizes any two mappings $h: X \rightarrow R$ and $i: Y \rightarrow R$, such that any mapping $i: Y \rightarrow R$, so that $h = r \circ f$, and furthermore $i = r \circ g$. Co-products are themselves special cases of the more general notion of co-limits. Figure~\ref{pullback}  illustrates the fundamental property of a {\em {pullback}}, which along with {\em pushforward}, is one of the core ideas in category theory. The pullback square with the objects $U,X, Y$ and $Z$ implies that the composite mappings $g \circ f'$ must equal $g' \circ f$. In this example, the morphisms $f$ and $g$ represent a {\em {pullback}} pair, as they share a common co-domain $Z$. The pair of morphisms $f', g'$ emanating from $U$ define a {\em {cone}}, because the pullback square ``commutes'' appropriately. Thus, the pullback of the pair of morphisms $f, g$ with the common co-domain $Z$ is the pair of morphisms $f', g'$ with common domain $U$. Furthermore, to satisfy the universal property, given another pair of morphisms $x, y$ with common domain $T$, there must exist another morphism $k: T \rightarrow U$ that ``factorizes'' $x, y$ appropriately, so that the composite morphisms $f' \ k = y$ and $g' \ k = x$. Here, $T$ and $U$ are referred to as {\em cones}, where $U$ is the limit of the set of all cones ``above'' $Z$. If we reverse arrow directions appropriately, we get the corresponding notion of pushforward. So, in this example, the pair of morphisms $f', g'$ that share a common domain represent a pushforward pair. 
For any set-valued functor $\delta: S \rightarrow$ {\bf {Sets}}, the Grothendieck category of elements $\int \delta$ can be shown to be a pullback in the diagram of categories. Here, {${\bf Set}_*$} is the category of pointed sets, and $\pi$ is a projection that sends a pointed set $(X, x \in X)$ to its \mbox{underlying set $X$.} 
\begin{figure}[h]
\centering
\begin{tikzcd}
  T
  \arrow[drr, bend left, "x"]
  \arrow[ddr, bend right, "y"]
  \arrow[dr, dotted, "k" description] & & \\
    & 
    U\arrow[r, "g'"] \arrow[d, "f'"]
      & X \arrow[d, "f"] \\
& Y \arrow[r, "g"] &Z
\end{tikzcd}
\begin{tikzcd}
  T
  \arrow[drr, bend left, "x"]
  \arrow[ddr, bend right, "y"]
  \arrow[dr, dotted, "k" description] & & \\
    & 
    \int \delta \arrow[r, "\delta'"] \arrow[d, "\pi_\delta"]
      & {\bf Set}_* \arrow[d, "\pi"] \\
& S \arrow[r, "\delta"] & {\bf Set}
\end{tikzcd}
\caption{(\textbf{Left})
 Universal Property of pullback mappings. (\textbf{Right}) The Grothendieck category of elements $\int \delta$ of any set-valued functor $\delta: S \rightarrow$ {\bf {Set}} can be described as a pullback in the diagram of categories. Here, {\bf Set}$_*$ is the category of pointed sets $(X, x \in X)$, and $\pi$ is the ``forgetful" functor that sends a pointed set $(X, x \in X)$ into the underlying set $X$.  } 
\label{pullback3}
\end{figure}
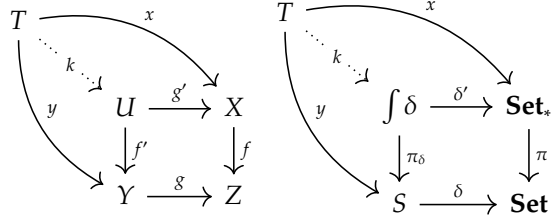

We can now proceed to define limits and colimits more generally. We define a {\em diagram} $F$ of {\em shape} $J$ in a category $C$ formally as a functor $F: J \rightarrow C$. We want to define the somewhat abstract concepts of {\em limits} and {\em colimits}, which will play a central role in this paper in defining a topos.  A convenient way to introduce these concepts is through the use of {\em universal cones} that are {\em over} and {\em under} a diagram.

For any object $c \in C$ and any category $J$, the {\em constant functor} $c: J \rightarrow C$ maps every object $j$ of $J$ to $c$ and every morphism $f$ in $J$ to the identity morphisms $1_c$. We can define a constant functor embedding as the collection of constant functors $\Delta: C \rightarrow C^J$ that send each object $c$ in $C$ to the constant functor at $c$ and each morphism $f: c \rightarrow c'$ to the constant natural transformation, that is, the natural transformation whose every component is defined to be the morphism $f$. 

\begin{definition}
    A {\bf cone over} a diagram $F: J \rightarrow C$ with the {\bf summit} or {\bf apex} $c \in C$ is a natural transformation $\lambda: c \Rightarrow F$ whose domain is the constant functor at $c$. The components $(\lambda_j: c \rightarrow Fj)_{j \in J}$ of the natural transformation can be viewed as its {\bf legs}. Dually, a {\bf cone under} $F$ with {\bf nadir} $c$ is a natural transformation $\lambda: F \Rightarrow c$ whose legs are the components $(\lambda_j: F_j \rightarrow c)_{j \in J}$.

\[\begin{tikzcd}
	&& c &&&& Fj &&&& Fk \\
	\\
	{F j} &&&& Fk &&&& c
	\arrow["{\lambda_j}", from=1-3, to=3-1]
	\arrow["{\lambda_k}"', from=1-3, to=3-5]
	\arrow["{F f}", from=3-1, to=3-5]
	\arrow["Ff", from=1-7, to=1-11]
	\arrow["{\lambda_j}"', from=1-7, to=3-9]
	\arrow["{\lambda_k}", from=1-11, to=3-9]
\end{tikzcd}\]
    
\end{definition}

Cones under a diagram are referred to usually as {\em cocones}. Using the concept of cones and cocones, we can now formally define the concept of limits and colimits more precisely. 

\begin{definition}
    For any diagram $F: J \rightarrow C$, there is a functor 

    \[ \mbox{Cone}(-, F): C^{op} \rightarrow \mbox{{\bf Set}} \]

    which sends $c \in C$ to the set of cones over $F$ with apex $c$. Using the Yoneda Lemma, a {\bf limit} of $F$ is defined as an object $\lim F \in C$ together with a natural transformation $\lambda: \lim F \rightarrow F$, which can be called the {\bf universal cone} defining the natural isomorphism 

    \[ C(-, \lim F) \simeq \mbox{Cone}(-, F) \]

    Dually, for colimits, we can define a functor 

    \[ \mbox{Cone}(F, -): C \rightarrow \mbox{{\bf Set}} \]

    that maps object $c \in C$ to the set of cones under $F$ with nadir $c$. A {\bf colimit} of $F$ is a representation for $\mbox{Cone}(F, -)$. Once again, using the Yoneda Lemma, a colimit is defined by an object $\mbox{Colim} F \in C$ together with a natural transformation $\lambda: F \rightarrow \mbox{colim} F$, which defines the {\bf colimit cone} as the natural isomorphism 

    \[ C(\mbox{colim} F, -) \simeq \mbox{Cone}(F, -) \]
\end{definition}

Limit and colimits of diagrams over arbitrary categories can often be reduced to the case of their corresponding diagram properties over sets. One important stepping stone is to understand how functors interact with limits and colimits. 

\begin{definition}
    For any class of diagrams $K: J \rightarrow C$, a functor $F: C \rightarrow D$ 

    \begin{itemize}
        \item {\bf preserves} limits if for any diagram $K: J \rightarrow C$ and limit cone over $K$, the image of the cone defines a limit cone over the composite diagram $F K: J \rightarrow D$. 

        \item {\bf reflects} limits if for any cone over a diagram $K: J \rightarrow C$ whose image upon applying $F$ is a limit cone for the diagram $F K: J \rightarrow D$ is a limit cone over $K$

        \item {\bf creates} limits if whenever $FK : J \rightarrow D$ has a limit in $D$, there is some limit cone over $F K$ that can be lifted to a limit cone over $K$ and moreoever $F$ reflects the limits in the class of diagrams. 
    \end{itemize}
\end{definition}

To interpret these abstract definitions, it helps to concretize them in terms of a specific universal construction, like the pullback defined above $c' \rightarrow c \leftarrow c''$ in $C$. Specifically, for pullbacks: 

\begin{itemize} 

\item A functor $F$ {\bf preserves pullbacks} if whenever $p$ is the pullback of  $c' \rightarrow c \leftarrow c''$ in $C$, it follows that $Fp$ is the pullback of  $Fc' \rightarrow Fc \leftarrow Fc''$ in $D$.

\item A functor $F$ {\bf reflects  pullbacks}  if  $p$ is the pullback of  $c' \rightarrow c \leftarrow c''$ in $C$ whenever $Fp$ is the pullback of  $Fc' \rightarrow Fc \leftarrow Fc''$ in $D$.

\item A functor $F$ {\bf creates pullbacks} if there exists some $p$ that is the pullback of  $c' \rightarrow c \leftarrow c''$ in $C$ whenever there exists a $d$ such  that $d$ is the pullback of  $Fc' \rightarrow Fc \leftarrow Fc''$ in $F$.

\end{itemize}

\subsection{Symmetric Monoidal Categories}
\label{smc}

Categorical models of causality \citep{fong:ms,fritz:jmlr,string-diagram-surgery,DBLP:journals/entropy/Mahadevan23} are usually defined over symmetric monoidal categories, which we briefly review now \citep{maclane:71,richter2020categories}.  

\begin{definition}
    A {\bf monoidal category} is a category {\cal C} together with a functor $\otimes: {\cal C} \times {\cal C} \rightarrow {\cal C}$, an identity object $e$ of {\cal C} and natural isomorphisms $\alpha, \lambda, \rho$ defined as: 

    \begin{eqnarray*}
        \alpha_{C_1, C_2, C_3}: C_1 \otimes (C_2 \otimes C_3) & \cong & (C_1 \otimes C_2) \otimes C_2,\\
        \lambda_C: e \otimes C & \cong & C,   \\
        \rho: C \otimes e & \cong & C, 
    \end{eqnarray*}
\end{definition}

The natural isomorphisms must satisfy coherence conditions called the ``pentagon" and ``triangle" diagrams \citep{maclane:71}. An important result shown in \citep{maclane:71} is that these coherence conditions guarantee that all well-formed diagrams must commute.  There are many natural examples of monoidal categories, the simplest one being the category of finite sets, termed {\bf FinSet} in \citep{Fritz_2020}, where each object $C$ is a set, and the tensor product $\otimes$ is the Cartesian product of sets, with functions acting as arrows. Deterministic causal models can be formulated in the category {\bf FinSet}. Other examples include the category of sets with relations as morphisms, and the category of Hilbert spaces \citep{Heunen2019}. Markov categories \citep{fritz:jmlr} are monoidal categories, where the identity element $e$ is also a terminal object, meaning there is a unique ``delete" morphism $d_e: X \rightarrow e$ associated with each object $X$. \citep{Fritz_2020} shows they form a unifying foundation for probabilistic and statistical reasoning. 
\begin{definition}
    A {\bf symmetric monoidal category} is a monoidal category $({\cal C}, \otimes, e, \alpha, \lambda, \rho)$ together with a natural isomorphism 
    \begin{eqnarray*}
       \tau_{C_1, C_2}: C_1 \otimes C_2 \cong C_2 \otimes C_1, \ \ \mbox{for all objects} \ \ C_1, C_2
    \end{eqnarray*}
    where $\tau$ satisfies the additional conditions: for all objects $C_1, C_2$ $\tau_{C_2, C_1} \circ \tau_{C_1, C_2} \cong 1_{C_1 \otimes C_2}$, and for all objects $C$, $\rho_C = \lambda_C \circ \tau_{C, e}: C \otimes e \cong C$. 
\end{definition}
An additional hexagon axiom is required to ensure that the $\tau$ natural isomorphism is compatible with $\alpha$.  The $\tau$ operator is called a ``swap" in Markov categories \citep{Fritz_2020}.  In most cases of interest in AI, the symmetric monoidal categories are {\em enriched} over some convenient base category ${\cal V}$, including vector spaces, or preorders such as the unit interval $[0,1]$, where the unique morphism from $a \rightarrow b$ exists if and only if $a \leq b$.
5
\begin{definition}
A {\bf {\cal V}-enriched category} consists of a regular category ${\cal C}$, such that for each pair of objects $x$ and $y$ in ${\cal C}$, the morphisms ${\cal C}(x,y) \in {\cal V}$, often referred to as a ${\cal V}$-hom object. For the case when $({cal V}, \leq, \otimes, 1)$ is a commutative monoidal preorder, we have the following conditions 
\begin{itemize}
    \item $1 \leq {\cal C}(x, x)$
    \item ${\cal C}(y,z) \otimes {\cal C}(x,y) \leq {\cal C}(x,z)$
\end{itemize}
\end{definition}

\subsection{Affine CDU and Markov Categories} 

\label{mccat}

In this section, we  review previous work on affine CDU (``copy-delete-uniform") categories \citep{Cho_2019} and Markov categories \citep{Fritz_2020}, which have been proposed as a unifying categorical model for causal inference, probability and statistics. They are symmetric monoidal categories, which we reviewed above, combined with a comonoidal structure on each object. Importantly, Markov categories are semi-Cartesian because they do not use uniform copying, but contain a Cartesian subcategory defined by deterministic morphisms (see below).  We give a brief review of Markov categories, and significant additional details that are omitted can be found in \citep{Cho_2019,Fritz_2020,fritz:jmlr}.  The equations are written in the diagrammatic language of string diagrams, which can be shown to represent a formal language that is equivalent to writing down algebraic equations \citep{Selinger_2010}. 

\begin{definition}
	\label{markov_cat}
	A \emph{Markov category} ${\cal C}$ \citep{Fritz_2020} is a symmetric monoidal category in which every object $X \in {\cal C}$ is equipped with a commutative comonoid structure given by a comultiplication $\cop_X : X \to X \otimes X$ and a counit $\del_X : X \to I$, depicted in string diagrams as
	\beq
		\label{cd}
		\begin{tikzpicture}
	\begin{pgfonlayer}{nodelayer}
		\node [style=bn] (0) at (7, 1) {};
		\node [style=bn] (1) at (-4, 0) {};
		\node [style=none] (2) at (7, -1) {};
		\node [style=none] (3) at (-4, -1) {};
		\node [style=none] (4) at (-5, 1) {};
		\node [style=none] (5) at (-3, 1) {};
		\node [style=none] (10) at (5, 0) {$=$};
		\node [style=none] (11) at (-7, 0) {$=$};
		\node [style=none] (12) at (3, 0) {$\del_X$};
		\node [style=none] (13) at (-9, 0) {$\cop_X$};
	\end{pgfonlayer}
	\begin{pgfonlayer}{edgelayer}
		\draw [bend right=45, looseness=1.25] (4.center) to (1);
		\draw (1) to (3.center);
		\draw [bend right=45] (1) to (5.center);
		\draw (2.center) to (0);
	\end{pgfonlayer}
\end{tikzpicture}

	\eeq
	and satisfying the commutative comonoid equations,
	\beq
		\label{comonoid_ass}
		\begin{tikzpicture}
	\begin{pgfonlayer}{nodelayer}
		\node [style=none] (4) at (-3.25, -2) {};
		\node [style=none] (5) at (2.25, 0) {};
		\node [style=bn] (8) at (-3.25, -1) {};
		\node [style=none] (11) at (4.25, 2) {};
		\node [style=bn] (14) at (-2.25, 1) {};
		\node [style=none] (17) at (-4.25, 2) {};
		\node [style=none] (18) at (3.25, 2) {};
		\node [style=none] (20) at (-4.25, 0) {};
		\node [style=none] (22) at (2.25, 0) {};
		\node [style=none] (23) at (-3.25, 2) {};
		\node [style=none] (24) at (3.25, -2) {};
		\node [style=none] (28) at (-2.25, 0) {};
		\node [style=none] (30) at (4.25, 0) {};
		\node [style=none] (31) at (0, 0) {$=$};
		\node [style=bn] (32) at (2.25, 1) {};
		\node [style=none] (33) at (-2.25, 0) {};
		\node [style=bn] (34) at (3.25, -1) {};
		\node [style=none] (35) at (1.25, 2) {};
		\node [style=none] (36) at (-1.25, 2) {};
	\end{pgfonlayer}
	\begin{pgfonlayer}{edgelayer}
		\draw [style=none] (4.center) to (8);
		\draw [style=none, bend left=45] (8) to (20.center);
		\draw [style=none, bend right=45] (8) to (28.center);
		\draw [style=none] (33.center) to (14);
		\draw [style=none, bend left=45] (14) to (23.center);
		\draw [style=none, bend right=45] (14) to (36.center);
		\draw [style=none] (20.center) to (17.center);
		\draw [style=none] (24.center) to (34);
		\draw [style=none, bend left=45] (34) to (5.center);
		\draw [style=none, bend right=45] (34) to (30.center);
		\draw [style=none] (22.center) to (32);
		\draw [style=none, bend left=45] (32) to (35.center);
		\draw [style=none, bend right=45] (32) to (18.center);
		\draw [style=none] (30.center) to (11.center);
	\end{pgfonlayer}
\end{tikzpicture}

	\eeq
	\beq
		\label{comonoid_other}
		\begin{tikzpicture}
	\begin{pgfonlayer}{nodelayer}
		\node [style=none] (0) at (-10, 1) {};
		\node [style=none] (1) at (-10, 2) {};
		\node [style=none] (2) at (-12, 1) {};
		\node [style=none] (3) at (-6.25, 0.5) {$=$};
		\node [style=none] (4) at (-3, 1) {};
		\node [style=none] (5) at (-11, -1) {};
		\node [style=bn] (6) at (-12, 1) {};
		\node [style=bn] (7) at (-4, 0) {};
		\node [style=none] (8) at (-5, 1) {};
		\node [style=bn] (9) at (-11, 0) {};
		\node [style=none] (10) at (-4, -1) {};
		\node [style=none] (11) at (-12, 1) {};
		\node [style=none] (12) at (-3, 1) {};
		\node [style=none] (13) at (-5, 2) {};
		\node [style=none] (14) at (-7.5, 2) {};
		\node [style=bn] (15) at (-3, 1) {};
		\node [style=none] (16) at (-8.75, 0.5) {$=$};
		\node [style=none] (17) at (-7.5, -1) {};
		\node [style=bn] (18) at (3.75, -0.25) {};
		\node [style=none] (19) at (2.75, 0.75) {};
		\node [style=none] (20) at (4.75, 0.75) {};
		\node [style=bn] (21) at (8.25, -0.25) {};
		\node [style=none] (22) at (7.25, 0.75) {};
		\node [style=none] (23) at (9.25, 0.75) {};
		\node [style=none] (24) at (6, 0) {$=$};
		\node [style=none] (25) at (3.75, -1.75) {};
		\node [style=none] (26) at (8.25, -1.75) {};
		\node [style=none] (27) at (2.75, 1.75) {};
		\node [style=none] (28) at (4.75, 1.75) {};
		\node [style=none] (29) at (7.25, 1.75) {};
		\node [style=none] (30) at (9.25, 1.75) {};
	\end{pgfonlayer}
	\begin{pgfonlayer}{edgelayer}
		\draw [style=none] (5.center) to (9);
		\draw [style=none, bend left=45] (9) to (11.center);
		\draw [style=none, bend right=45] (9) to (0.center);
		\draw [style=none] (6) to (2.center);
		\draw [style=none] (0.center) to (1.center);
		\draw [style=none] (10.center) to (7);
		\draw [style=none, bend left=45] (7) to (8.center);
		\draw [style=none, bend right=45] (7) to (12.center);
		\draw [style=none] (8.center) to (13.center);
		\draw [style=none] (15) to (4.center);
		\draw [style=none] (17.center) to (14.center);
		\draw [style=none, bend left=45] (18) to (19.center);
		\draw [style=none, bend right=45] (18) to (20.center);
		\draw [style=none, bend left=45] (21) to (22.center);
		\draw [style=none, bend right=45] (21) to (23.center);
		\draw (26.center) to (21);
		\draw (25.center) to (18);
		\draw (30.center) to (23.center);
		\draw (29.center) to (22.center);
		\draw [in=90, out=-90, looseness=0.75] (27.center) to (20.center);
		\draw [in=-90, out=90, looseness=0.75] (19.center) to (28.center);
	\end{pgfonlayer}
\end{tikzpicture}

	\eeq
	as well as compatibility with the monoidal structure,
	\begin{equation}
		\label{delcopyAB}
		\resizebox{0.8\textwidth}{0.1\textwidth}{%
\begin{tikzpicture}
	\begin{pgfonlayer}{nodelayer}
		\node [style=none] (0) at (-7, 0) {$=$};
		\node [style=none] (1) at (2, 2.5) {$X\otimes Y$};
		\node [style=none] (2) at (11, -2) {};
		\node [style=bn] (3) at (11, -1) {};
		\node [style=none] (4) at (12, 0) {};
		\node [style=none] (5) at (12, 0) {};
		\node [style=none] (6) at (10, 0) {};
		\node [style=none] (7) at (15, -2) {};
		\node [style=bn] (8) at (15, -1) {};
		\node [style=none] (9) at (16, 0) {};
		\node [style=none] (10) at (16, 0) {};
		\node [style=none] (11) at (14, 0) {};
		\node [style=none] (12) at (10, 2) {};
		\node [style=none] (13) at (12, 2) {};
		\node [style=none] (14) at (14, 2) {};
		\node [style=none] (15) at (16, 2) {};
		\node [style=none] (16) at (-9, -1.5) {$X\otimes Y$};
		\node [style=none] (17) at (-5, -1.5) {$X$};
		\node [style=none] (18) at (8, 0) {$=$};
		\node [style=bn] (19) at (-9, 1) {};
		\node [style=bn] (20) at (-5, 1) {};
		\node [style=bn] (21) at (-3, 1) {};
		\node [style=bn] (22) at (4.5, -0.5) {};
		\node [style=none] (23) at (4.5, -2) {};
		\node [style=none] (24) at (3, 1) {};
		\node [style=none] (25) at (6, 1) {};
		\node [style=none] (26) at (-5, -1) {};
		\node [style=none] (27) at (-3, -1) {};
		\node [style=none] (28) at (-3, -1.5) {$Y$};
		\node [style=none] (29) at (-9, -1) {};
		\node [style=none] (30) at (6, 2.5) {$X\otimes Y$};
		\node [style=none] (31) at (4.5, -2.5) {$X\otimes Y$};
		\node [style=none] (32) at (3, 2) {};
		\node [style=none] (33) at (6, 2) {};
		\node [style=none] (34) at (10, 2.5) {$X$};
		\node [style=none] (35) at (12, 2.5) {$Y$};
		\node [style=none] (36) at (14, 2.5) {$X$};
		\node [style=none] (37) at (16, 2.5) {$Y$};
		\node [style=none] (38) at (11, -2.5) {$X$};
		\node [style=none] (39) at (15, -2.5) {$Y$};
	\end{pgfonlayer}
	\begin{pgfonlayer}{edgelayer}
		\draw [style=none] (2.center) to (3);
		\draw [style=none, bend left=45] (3) to (6.center);
		\draw [style=none, bend right=45] (3) to (5.center);
		\draw [style=none] (7.center) to (8);
		\draw [style=none, bend left=45] (8) to (11.center);
		\draw [style=none, bend right=45] (8) to (10.center);
		\draw [style=none] (12.center) to (6.center);
		\draw [style=none] (9.center) to (15.center);
		\draw [style=none, in=-90, out=90] (4.center) to (14.center);
		\draw [style=none, in=90, out=-90] (13.center) to (11.center);
		\draw (29.center) to (19);
		\draw (26.center) to (20);
		\draw (27.center) to (21);
		\draw (23.center) to (22);
		\draw [bend right=315] (22) to (24.center);
		\draw [bend right=45] (22) to (25.center);
		\draw (33.center) to (25.center);
		\draw (24.center) to (32.center);
	\end{pgfonlayer}
\end{tikzpicture}
}

	\end{equation}
	and naturality of $\del$, which means that
	\begin{equation}
		\label{counit_nat}
		\begin{tikzpicture}
	\begin{pgfonlayer}{nodelayer}
		\node [style=bn] (0) at (-1.5, 2.5) {};
		\node [style=none] (8) at (-1.5, -1.5) {};
		\node [style=bn] (9) at (1.5, 1.5) {};
		\node [style=none] (10) at (1.5, -1.5) {};
		\node [style=none] (11) at (0.5, 0) {$=$};
		\node [style=morphism] (12) at (-1.5, 0) {$f$};
	\end{pgfonlayer}
	\begin{pgfonlayer}{edgelayer}
		\draw [style=none] (9) to (10.center);
		\draw (8.center) to (12);
		\draw (12) to (0);
	\end{pgfonlayer}
\end{tikzpicture}

	\end{equation}
	for every morphism $f$.
\end{definition}

Let us briefly explain these definitions. The $\mbox{del}_X: X \rightarrow I$ is essentially like integrating over a probability distribution, which always yields $1$. Hence, $I$, the unit of the tensor product, is the terminal object in affine CDU and Markov categories. Bayes rule turns into a {\em disintegration rule} \citep{Cho_2019}, which is only available in Markov categories with conditionals (i.e., where one can categorically refine $P(y | x)$ conditional distributions). Note that in the continuous case of random variables defined as measurable functions on real numbers, one has to take considerable care in defining conditioning \citep{halmos:book}. The $\mbox{copy}_X$ procedure is uniform, and deterministic, meaning if you take the tensor product of two variables $X \otimes Y$ and then copy the resulting object, that's exactly the same as first copying $X$ into $X \otimes X$ and $Y$ into $Y \otimes Y$, and then taking the tensor product, along with a swap operation (see Equation~\ref{delcopyAB}). Only $\mbox{del}_X$ acts ``uniformly", meaning that if you process a variable $X$ using some function $f$ and then delete $f(X)$ (meaning marginalize it), that's equivalent to simply deleting $X$. However, $\mbox{copy}_X$ is not defined this way, and we discuss that subtlety below, as it will be important in understanding why Markov categories are semi-Cartesian. To convert them into a topos, we need the result to be Cartesian closed, which is why we need to use the Yoneda Lemma to construct the category of presheaves to guarantee obtaining a topos. 

\subsection{Cartesian Structure in Markov Categories}

\label{fox}

We now discuss a subcategory of Cartesian categories within  Markov that involves uniform $\mbox{copy}_X$ and $\mbox{del}_X$ morphisms. One fundamental property of Markov categories is that they are {\em semi-Cartesian}, as the unit object is also a terminal object.  But, a subtlety arises in how these copy and delete operators are modeled, as we discuss below. 

\begin{definition}
    A symmetric monoidal category ${\cal C}$ is {\bf Cartesian} if the tensor product $\otimes$ is the categorical product. 
\end{definition}

If ${\cal C}$ and ${\cal D}$ are symmetric monoidal categories, then a functor $F: {\cal C} \rightarrow {\cal D}$ is monoidal if the tensor product is preserved up to coherent natural isomorphisms. $F$ is strictly monoidal if all the monoidal structures are preserved exactly, including $\otimes$, unit object $I$, symmetry, associative and unit natural isomorphisms. Denote the category of symmetric monoidal categories with strict functors as arrows as {\bf MON}. Let us review the basic definitions given by \citet{Heunen2019}, which will give some further clarity on the Cartesian structure in affine CDU and Markov categories. 

\begin{definition}
 The subcategory of comonoids {\bf coMON} in the ambient category {\bf MON} of all symmetric monoidal categories is defined for any specific category {\cal C} as a collection of ``coalgebraic" objects $(X, \mbox{copy}_X, \mbox{del}_X)$, where $X$ is in {\cal C}, and arrows defined as comonoid homomorphisms  from $(X, \mbox{copy}_X, \mbox{del}_X)$ to  $(Y, \mbox{copy}_Y, \mbox{del}_Y)$ that act uniformly, in the sense that if $f: X \rightarrow Y$ is any morphism in {\cal C}, then: 

 \begin{eqnarray*}
     (f \otimes f) \circ \mbox{copy}_X = \mbox{copy}_Y \circ f \\
     \mbox{del}_Y \circ f = \mbox{del}_X 
 \end{eqnarray*}
\end{definition}

\citet{Heunen2019} define the process of ``uniform copying and deleting" in the category {\bf coMON}, which we now relate to Markov categories. A subtle difference worth emphasizing with Definition~\ref{markov_cat} is that in Markov categories, only $\mbox{del}_X$ is ``uniform", but not $\mbox{copy}_X$ in the sense defined by \citet{Heunen2019}.  This distinction can be modeled in a cPROP category that is semi-Cartesian like Markov categories by suitably modifying the definition of the associated PROP map for copying. 

\begin{definition} \citep{Heunen2019}
    A symmetric monoidal category {\cal C} admits {\bf uniform deleting} if there is a natural transformation $e_X: X \xrightarrow{e_X} I$ for all objects in the  subcategory ${\cal C}_{\bf coMON}$ of comonoidal objects, where $e_I = \mbox{id}_I$, as shown in Equation~\ref{counit_nat}. 
\end{definition}

This condition was referred to by \citet{Cho_2019} as a {\em causality} condition on the arrow $e_X$.  Essentially, it states that if you process some object and then discard it, it's equivalent to discarding it without processing. 

\begin{theorem} \citep{Heunen2019}
A symmetric monoidal category {\cal C} has uniform deleting if and only if $I$ is terminal. 
\end{theorem}

This property holds for Markov categories, as noted in \citep{Fritz_2020}, and a simple diagram chasing proof is given in \citep{Heunen2019}. 

\begin{definition}\citep{Heunen2019}
    A symmetric monoidal category {\cal C} has {\bf uniform copying} if there is a natural transformation $\mbox{copy}_X: X \rightarrow X \otimes X$ such that $\mbox{del}_I = \rho^{-1}_I$ satisfying Equation~\ref{comonoid_ass} and Equation~\ref{comonoid_other}. 
\end{definition}

We can now state an important result proved in \citep{Heunen2019} (Theorem 4.28), which relates to the more general results  shown earlier by \citet{fox}.  

\begin{theorem}\citep{fox,Heunen2019}
    The following conditions are equivalent for a symmetric monoidal category {\cal C}. 

    \begin{itemize} 

    \item The category {\cal C} is {\bf Cartesian} with tensor products $\otimes$ given by the categorical product and the tensor unit is given by the terminal object. 

    \item The symmetric monoidal category {\cal C} has {\bf uniform copying and deleting}, and Equation~\ref{comonoid_ass} holds. 

    \end{itemize} 
\end{theorem}

As noted by \citet{Fritz_2020}, not all Markov categories are Cartesian, because their {\bf copy}$_X$ is not uniform, but only {\bf del}$_X$ is. For example, consider the  category {\bf FinStoch}, where a joint distribution is specified by the morphism $\psi: I \rightarrow X \otimes Y$. In this case, the marginal distributions can be formed as the composite morphisms
\begin{eqnarray*}
    I \xrightarrow{\psi} X \otimes  Y \xrightarrow{\mbox{del}_Y} X \\ 
    I \xrightarrow{\psi} X \otimes  Y \xrightarrow{\mbox{del}_X} Y \\ 
\end{eqnarray*}
But to require that in this case $\otimes$ is the categorical product implies that the marginal distributions defined as the above composites must be in bijection with the joint distribution.

\section{Topos Theory Background}

\subsection{Topos Category}

To help build intuition for a topos, let's understand why sets are special as a category. They have all limits and colimits, meaning that one can always construct the categorical product of two sets as the Cartesian product, and the disjoint union as the coproduct. They have all pullbacks, which are commutative diagrams that define in essence a categorical definition of products and coproducts. In terms of the Yoneda Lemma, a universal property is either an initial or final property in a category of diagrams (functors). The product is the final object in a category of diagrams, and the coproduct is the initial object. Every concept in category theory essentially reduces down to these simple notions. Sets also have exponential objects: the set of all functions between two sets is once again a set! They also have a subobject classifier: each set has subsets, defining its parts, which correspond to a characteristic function that evaluates to true for elements in the subset. Topos theory generalizes all these properties to categories. It is important to causal inference because a causal intervention defines a subobject of an arbitrary object. A causal intervention on an SCM produces a subobject of that SCM object. 

To define a topos, we need to go through a few more definitions.  

\begin{definition}
    An object $x$  in a category ${\cal C}$ is called {\bf initial} if there is a unique morphism $f: x \rightarrow y$ to every other object $y$ in the category. Dually, an object is called {\bf final} if there is a unique morphism $f: y \rightarrow x$ into $x$. 
\end{definition}

In the category ${\bf Sets}$, the null or empty set $\emptyset$ is the initial and the single element set $\{* \}$ is the final object. From an empty set, there is only one function possible to any other set, namely the empty function. From any set there is exactly one function into the single element set. A topos generalizes the property of subobject classifiers in {\bf Sets}. Given any subset $S \subset X$, we can define $S$ as the monic arrow $S \hookrightarrow X$ defined by the inclusion of $S$ in $X$, or as the characteristic function $\phi_S$ that is equal to $1$ for all elements $x \in X$ that belong to $S$, and takes the value $0$ otherwise. We can define the set ${\bf 2} = \{0, 1 \}$ and treat {\bf true} as the inclusion $\{1 \}$ in ${\bf 2}$. The characteristic function $\phi_S$ can then be defined as the pullback of {\bf true} along $\phi_S$. 

\[\begin{tikzcd}
	S &&& {{\bf 1}} \\
	\\
	X &&& {{\bf 2}}
	\arrow["m", tail, from=1-1, to=3-1]
	\arrow[from=1-1, to=1-4]
	\arrow["{{\bf true}}"{description}, tail, from=1-4, to=3-4]
	\arrow["{\phi_S}"{description}, dashed, from=3-1, to=3-4]
\end{tikzcd}\]

We can now define subobject classifiers in a category ${\cal C}$ as follows. 

\begin{definition}
    In a category ${\cal C}$ with finite limits, a {\bf subobject classifier} is a {\em monic} arrow ${\bf true}: {\bf 1} \rightarrow \Omega$, such that to every other monic arrow $S \hookrightarrow X$ in ${\cal C}$, there is a unique arrow $\phi$ that forms the following pullback square: 

\[\begin{tikzcd}
	S &&& {{\bf 1}} \\
	\\
	X &&& \Omega
	\arrow["m", tail, from=1-1, to=3-1]
	\arrow[from=1-1, to=1-4]
	\arrow["{{\bf true}}"{description}, tail, from=1-4, to=3-4]
	\arrow["{\phi}"{description}, dashed, from=3-1, to=3-4]
\end{tikzcd}\]
    
\end{definition}

This definition can be rephrased as saying that the subobject functor is representable. In other words, a subobject of an object $x$ in a category ${\cal C}$ is an equivalence class of monic arrows $m: S \hookrightarrow  x$. 

\begin{definition}
\label{toposdefn}
    A {\bf topos} is a category ${\cal E}$ with 
    \begin{enumerate}
        \item A pullback for every diagram $X \rightarrow B \leftarrow Y$. 

        \item A terminal object ${\bf 1}$. 

        \item An object $\Omega$ and a monic arrow ${\bf true}: 1 \rightarrow \Omega$ such that any monic $m: S \hookrightarrow B$, there is a unique arrow $\phi: B \rightarrow \Omega$ in ${\cal E}$ for which the following square is a pullback: 
        
        \[\begin{tikzcd}
	S &&& {{\bf 1}} \\
	\\
	X &&& \Omega
	\arrow["m", tail, from=1-1, to=3-1]
	\arrow[from=1-1, to=1-4]
	\arrow["{{\bf true}}"{description}, tail, from=1-4, to=3-4]
	\arrow["{\phi}"{description}, dashed, from=3-1, to=3-4]
\end{tikzcd}\]

\item To each object $x$ an object $P x$ and an arrow $\epsilon_x: x \times P x \rightarrow \Omega$ such that for every arrow $f: x \times y \rightarrow \Omega$, there is a unique arrow $g: y \rightarrow P x$ for which the following diagrams commute: 

\[\begin{tikzcd}
	y && {x \times y} &&& \Omega \\
	\\
	Px && {x \times P x} &&& \Omega
	\arrow["g", dashed, from=1-1, to=3-1]
	\arrow["f", from=1-3, to=1-6]
	\arrow["{\epsilon_x}", from=3-3, to=3-6]
	\arrow["{1 \times g}"{description}, dashed, from=1-3, to=3-3]
\end{tikzcd}\]

    \end{enumerate}
\end{definition}

Let us understand these definitions in the category of {\bf Sets}. Clearly, the single point set $\{ \bullet \}$ is a terminal object for {\bf Sets}, because there is a unique function from any set $S$ to a single element set $\bullet$, and the categorical product of two sets $A \times B$ is just the Cartesian product. Furthermore, given two sets $A$ and $B$, we can define $B^A$ as the exponential object representing the set of all functions $f: A \rightarrow B$. We can define exponential objects in any category more generally as follows. 

\begin{definition}
    Given any category ${\cal C}$ with products, for a fixed object $x$ in ${\cal C}$, we can define the functor 

    \[ x \times - :  \rightarrow {\cal C}\]

    If this functor has a right adjoint, which can be denoted as 

    \[ (-)^x: {\cal C} \rightarrow {\cal C} \]

    then we say $x$ is an {\bf exponentiable} object of ${\cal C}$. 
\end{definition}

\begin{definition}
    A category ${\cal C}$ is {\bf Cartesian closed} if it has finite products (which is equivalent to saying it has a terminal object and binary products) and if all objects in ${\cal C}$ are {\em exponentiable}. 
\end{definition}

\subsection{Sheaves and Toposes: Categories of Functors}

\label{toposes}

\begin{figure}[h] 
\centering
\caption{Two applications of sheaf theory in AI: (top) minimizing travel costs in weighted graphs satisfies the sheaf principle, one example of which is the Bellman optimality principle in reinforcement learning \citep{bertsekas:rlbook} (bottom): Approximating a function over a topological space must satisfy the sheaf condition. \label{sheaves}}
\vskip 0.1in
\begin{minipage}{0.7\textwidth}
\vskip 0.1in
\includegraphics[scale=0.35]{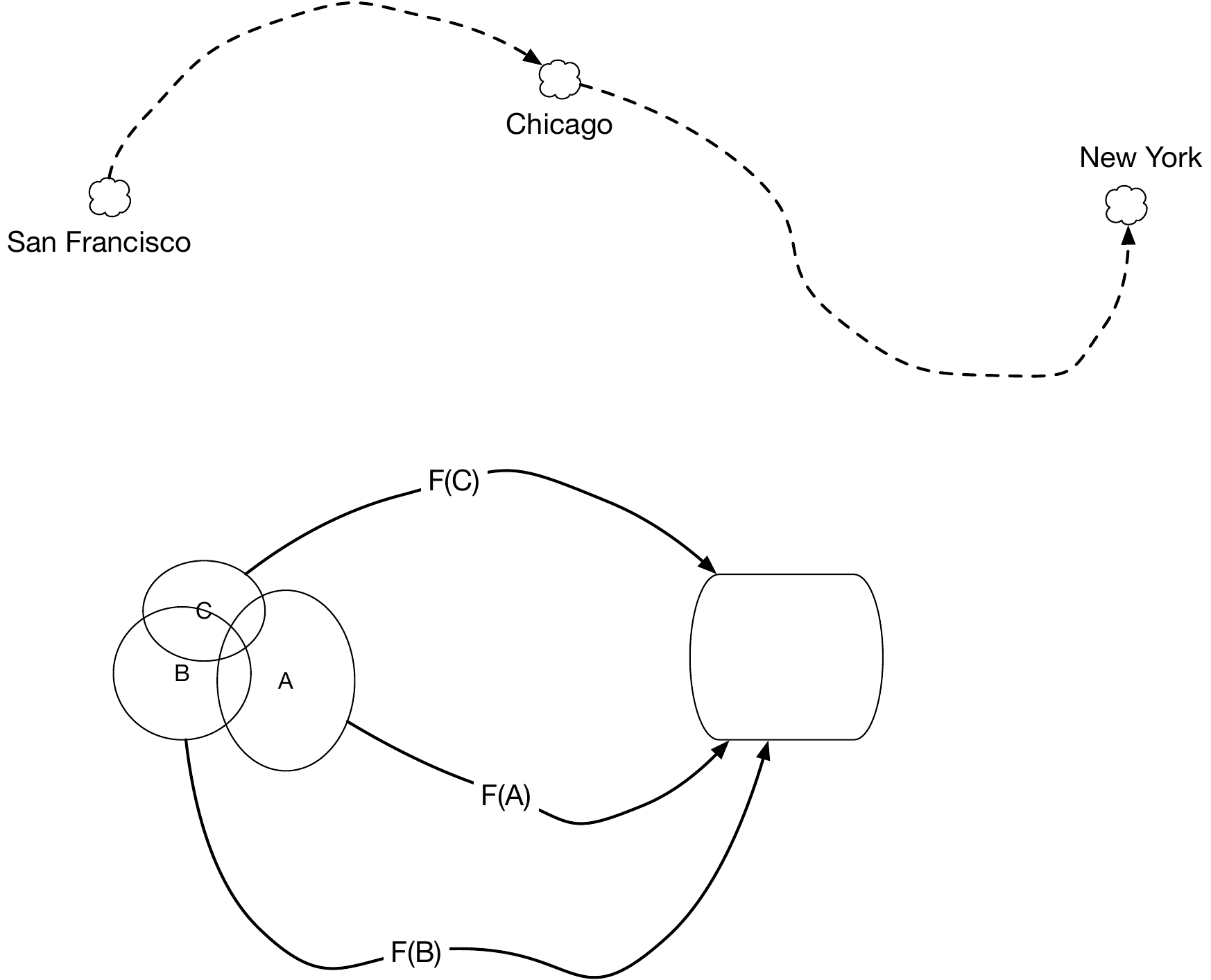}
\end{minipage}
\end{figure}

A general way to construct a topos category is through covariant Yoneda embeddings $\yo: {\cal C} \rightarrow {\bf Set}^{\cal C}$, or contravariant  Yoneda embeddings $\yo: {\cal C} \rightarrow {\bf Set}^{{\cal C}^{op}}$. In simpler terms, each object $x$ in the category ${\cal C}$ is either mapped to the functor ${\cal C}(x, -): {\cal C} \rightarrow {\bf Sets}$ or ${\cal C}(-, x): {\cal C}^{op} \rightarrow {\bf Sets}$. These structures are called presheaves or copresheaves. To construct a proper sheaf, we need to include an additional condition that is illustrated in Figure~\ref{sheaves}. The sheaf condition plays an important role in many applications of machine learning, from dimensionality reduction \citep{umap} to causal inference \citep{DBLP:journals/entropy/Mahadevan23}. \citet{maclane1992sheaves} provides an excellent overview of sheaves and topoi, and how remarkably they unify much of mathematics, from geometry to logic and topology.  For causal inference, the structure of causality shapes the arrows in a causal model, such as a Markov category, and that imposes a Grothendieck topology with an associated internal logic. Logic and causality are interwoven in ths sense. 

Figure~\ref{sheaves} gives two concrete examples of sheaves (in both cases, these are enriched sheaves).  In a minimum cost transportation problem, say reinforcement learning \citep{bertsekas:rlbook}, any optimal solution has the property that any restriction of the solution must also be optimal. In RL, this sheaf principle is codified by the Bellman equation, and leads to the fundamental principle of dynamic programming \citep{bertsekas:rlbook}. Consider routing candy bars from San Francisco to New York city. If the cheapest way to route candy bars is through Chicago, then the restriction of the overall route to the (sub) route from Chicago to New York City must also be optimal, otherwise it is possible to find a shortest overall route by switching to a lower cost route. Similarly, in function approximation with real-valued functions $F: {\cal C} \rightarrow \mathbb{R}$, where ${\cal C}$ is the category of topological spaces, the (sub)functions $F(A), F(B)$ and $F(C)$ restricted to the open sets $A$, $B$ and $C$ must agree on the values they map the elements in the intersections $A \cap B$, $A \cap C$, $A \cap B \cap C$ and so on. 

Sheaves can be defined over arbitrary categories, and we introduce the main idea by focusing on the category of sheaves over {\bf Sets}. 

\begin{definition}\citep{maclane1992sheaves}
    A {\bf sheaf} of sets $F$ on a topological space $X$ is a functor $F: {\cal O}^{op} \rightarrow {\bf Sets} $ such that each open covering $U = \bigcup_i U_i, i \in I$ of an open set $O$ of $X$ yields an equalizer diagram

    \[ FU \xrightarrow[]{e} \prod_i FU_i \doublerightarrow{p}{q} \prod_{i,j} F(U_i \cap U_j) \]
    
\end{definition}

The above definition succinctly generalizes the idea in SCMs of combining local functions to get a unique global function. 

\begin{definition}
    The category $\mbox{Sh}(X)$ of sheaves over a space $X$ is a full subcategory of the functor category ${\bf Sets}^{{\cal O}(X)^{op}}$.
\end{definition}

\subsection{Grothendieck Topologies}

We can generalize the notion of sheaves to arbitrary categories using the Yoneda embedding $\yo(x) = {\cal C}(-, x)$. We explain this generalization in the context of a more abstract topology on categories called the {\em Grothendieck topology} defined by {\em sieves}. A sieve can be viewed as a {\em subobject} $S \subseteq \yo(x)$ in the presheaf ${\bf Sets}^{{\cal C}^{op}}$, but we can define it more elegantly as a family of morphisms in ${\cal C}$, all with codomain $x$ such that

\[ f \in S \Longrightarrow f \circ g \in S \]

\begin{figure}[t] 
\centering
\caption{Sieves are subobjects of of $\yo(x)$ Yoneda embeddings of a category ${\cal C}$, which generalizes the concept of sheaves over sets in Figure~\ref{sheaves}. \label{sieves}}
\vskip 0.1in
\begin{minipage}{0.7\textwidth}
\vskip 0.1in
\includegraphics[scale=0.35]{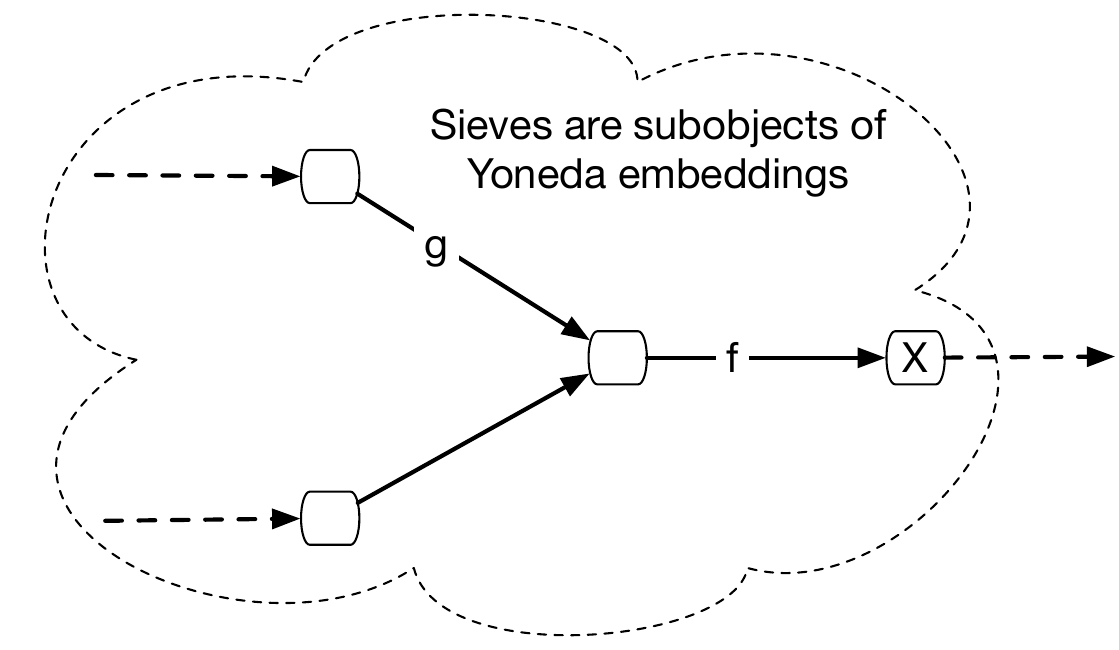}
\end{minipage}
\end{figure}

Figure~\ref{sieves} illustrates the idea of sieves. A simple way to think of a sieve is as a {\em right ideal}. We can define that more formally as follows: 

\begin{definition}
    If $S$ is a sieve on $x$, and $h: D \rightarrow x$ is any arrow in category ${\cal C}$, then 

    \[ h^*(S) = \{g \ | \ \mbox{cod}(g) = D, hg \in S \}\]
\end{definition}

\begin{definition}\citep{maclane1992sheaves}
A {\bf Grothendieck topology} on a category ${\cal C}$ is a function $J$ which assigns to each object $x$  of ${\cal C}$ a collection $J(x)$ of sieves on $x$ such that
\begin{enumerate}
    \item the maximum sieve $t_x = \{ f | \mbox{cod}(f) = x \}$ is in $J(x) $. 
    \item If $S \in J(x)$  then $h^*(S) \in J(D)$ for any arrow $h: D \rightarrow x$. 
    \item If $S \in J(x)$ and $R$ is any sieve on $x$, such that $h^*(R) \in J(D)$ for all $h: D \rightarrow x$, then $R \in J(C)$. 
\end{enumerate}
\end{definition}

We can now define categories with a given Grothendieck topology as {\em sites}. 

\begin{definition}
    A {\bf site} is defined as a pair $({\cal C}, J)$ consisting of a small category ${\cal C}$ and a Grothendieck topology $J$ on ${\cal C}$. 
\end{definition}

An intuitive way to interpret a site is as a generalization of the notion of a topology on a space $X$, which is defined as a set $X$ together with a collection of open sets ${\cal O}(X)$. The sieves on a category play the role of ``open sets".

\subsection{Mitchell-B\'enabou Language for TCMs}
\label{mbl}
We define the Mitchell-B\'enabou language (MBL), a typed local set theory (see Section~\ref{lst}) associated with a causal topos. Given the topos category ${\cal C}_{TCM}$, we define the types of MBL as causal model objects $M$ of ${\cal C}_{TCM}$. For each type $M$ (e.g., an SCM),  we assume the existence of variables $x_M, y_M, \ldots$, where each such variable has as its interpretation the identity arrow ${\bf 1}: M \rightarrow M$. We can construct product objects, such as $A \times B \times C$, where terms like $\sigma$ that define arrows are given the interpretation $\sigma: A \times B \times C \rightarrow D$. We can inductively define the terms and their interpretations in a topos category as follows (see \citep{maclane1992sheaves} for additional details): 
\begin{itemize}
    \item Each variable $x_M$ of type $M$ is a term of type $M$, and its interpretation is the identity $x_M = {\bf 1}: M \rightarrow M$ (e.g., $M$ may be an SCM or a causal model on a  Markov category). 

\item Terms $\sigma$ and $\tau$ of types $C$ and $D$ that are interpreted as $\sigma: A \rightarrow C$ and $\tau: B \rightarrow D$ can be combined to yield a term $\langle \sigma, \tau \rangle$ of type $C \times D$, whose joint interpretation is given as 

\[ \langle \sigma p, \tau q \rangle: X \rightarrow C \times D\]

where $X$ has the required projections $p: X \rightarrow A$  and $q: X \rightarrow B$. 

\item Terms $\sigma: A \rightarrow B$ and $\tau: C \rightarrow B$ of the same type $B$ yield a term $\sigma = \tau$ of type $\Omega$, interpreted as 

\[ (\sigma = \tau): W \xrightarrow[]{\langle \sigma p, \tau q \rangle} B \times B \xrightarrow[]{\delta_B} \Omega \]
where $\delta_B$ is the characteristic map of the diagonal functor $\Delta B \rightarrow B \times B$. These diagonal maps correspond to the ``copy" procedure in Markov categories \citep{Fritz_2020}. 

\item Arrows $f: A \rightarrow B$ and a term $\sigma: C \rightarrow A$ of type $A$ can be combined to yield a term $f \circ \sigma$ of type $B$, whose interpretation is naturally a composite arrow: 

\[ f \circ \sigma: C \xrightarrow[]{\sigma} A \xrightarrow[]{f} B\]

\item For exponential objects, terms $\theta: A \rightarrow B^C$ and $\sigma: D \rightarrow C$ of types $B^C$ and $C$, respectively, combine to give an ``evaluation" map of type $B$, defined as 

\[ \theta (\sigma): W \rightarrow B^C \times C \xrightarrow[]{e} B \]

where $e$ is the evaluation map, and $W$ defines a map $\langle \theta p, \sigma q \rangle$, where once again $p: W \rightarrow A$ and $q: W \rightarrow D$ are projection maps. 

\item Terms $\sigma: A \rightarrow B$ and $\tau: D \rightarrow \Omega^B$ combine to yield a term $\sigma \in \tau$ of type $\Omega$, with the following interpretation: 

\[ \sigma \in \tau: W \xrightarrow[]{\langle \sigma p, \tau q \rangle} B \times \Omega^B \xrightarrow[]{e} \Omega \]

\item Finally, we can define local functions as $\lambda$ objects, such as 

\[ \lambda x_C \sigma: A \rightarrow B^C \]

where $x_C$ is a variable of type $C$ and $\sigma: C \times A \rightarrow B$. 
\end{itemize}

We combine terms $\alpha, \beta$ etc. of type $\Omega$ using logical connectives $\wedge, \vee, \Rightarrow, \neg$, as well as quantifiers, to get composite terms, where each of the logical connectives is now defined over the subobject classifier $\Omega$. 


\begin{itemize}
    \item $\wedge: \Omega \times \Omega \rightarrow \Omega$ is interpreted as the {\em meet} operation in the partially ordered set of subobjects (given by the Heyting algebra). 

    \item $\vee: \Omega \times \Omega \rightarrow \Omega$ is interpreted as the {\em join} operation in the partially ordered set of subobjects (given by the Heyting algebra).  This operation gives the definition of a disjunction of two properties. 

    \item $\Rightarrow: \Omega \times \Omega \rightarrow \Omega$ is interpreted as an adjoint functor, as defined previously for a Heyting algebra. Thus, the property of implication over SCMs is modeled as an adjoint functor. 
    
\end{itemize}

We can combine these logical connectives with the term interpretation as arrows, relegating some details to \citep{maclane1992sheaves}. We now turn to the Kripke-Joyal semantics of this language. 

\subsection{Kripke-Joyal Semantics for TCMs}
\label{kj}

 We now define the Kripke-Joyal semantics for the Mitchell-B\'enabou language of a causal topos. Any free variable $x$ must have some causal model $X$ of ${\cal C}_{TCM}$ as its type.  For any causal model $M$ in ${\cal C}_{TCM}$, define a {\em generalized element} as a morphism $\alpha: N \rightarrow M$. To understand this definition, note that we can define an element of a causal model by the  morphism $x: {\bf 1} \rightarrow M$. Thus, a generalized element $\alpha: N \rightarrow M$ represents the ``stage of definition" of $M$ by $N$. We specify the semantics of how an SCM $N$ supports any formula $\phi(\alpha)$, denoted by $N \Vdash \phi(\alpha)$, as follows: 
\[ N \Vdash \phi(\alpha) \ \ \ \mbox{if and only if } \ \ \ \mbox{Im} \ \alpha \leq \{ x | \phi(x) \} \]
Stated in the form of a commutative diagram, this ``forcing" relationship holds if and only if $\alpha$ factors through $\{x | \phi(x) \}$, where $x$ is a variable of type $M$ (recall that objects $M$ of a topos form its types), as shown in the following commutative diagram. \footnote{The concept of ``forcing" is generalized from set theory \citep{maclane1992sheaves}.}
 \begin{center}
\begin{tikzcd}
	&& {\{x | \phi(x) \}} && {{\bf 1}} \\
	\\
	N && M && {{\bf \Omega} }
	\arrow[from=1-3, to=1-5]
	\arrow[tail, from=1-3, to=3-3]
	\arrow["{{\bf True}}", from=1-5, to=3-5]
	\arrow[dashed, from=3-1, to=1-3]
	\arrow["\alpha"', from=3-1, to=3-3]
	\arrow["{\phi(x)}", from=3-3, to=3-5]
\end{tikzcd}
 \end{center}
This diagram provides an interesting way to define causal interventions in a causal topos, because it defines submodels of $M$.   Building on this definition, if $\alpha, \beta: N \rightarrow M$ are parallel arrows, we can give semantics to the formula $\alpha = \beta$ by the following statement: 

 \[ N \xrightarrow[]{\langle \alpha, \beta \rangle} M \times M \xrightarrow[]{\delta_M} \Omega\]

following the definitions in the previous section for the composite $\langle \alpha, \beta \rangle$ and $\delta_X$ in the Mitchell-B\'enabou language. We can extend the previous commutative diagram to show that $U \Vdash \alpha = \beta$ holds if and only if $\langle \alpha, \beta \rangle$ factors through the diagonal map $\Delta$: 

\begin{center}
\begin{tikzcd}
	&& M && {{\bf 1}} \\
	\\
	N && {M \times M} && {{\bf \Omega} }
	\arrow[from=1-3, to=1-5]
	\arrow["\Delta", tail, from=1-3, to=3-3]
	\arrow["{{\bf True}}", from=1-5, to=3-5]
	\arrow[dashed, from=3-1, to=1-3]
	\arrow["{\langle \alpha, \beta \rangle}"', from=3-1, to=3-3]
	\arrow["{\delta_M}", from=3-3, to=3-5]
\end{tikzcd}
\end{center}


\begin{itemize}
    \item {\bf Monotonicity:} If $U \Vdash \phi(x)$, then we can pullback the interpretation through any arrow $f: U' \rightarrow U$ in a topos ${\cal C}$ to obtain $U' \Vdash \phi(\alpha \circ f)$. 
    \begin{small}
\begin{tikzcd}
	&&&& {\{x | \phi(x) \}} && {{\bf 1}} \\
	\\
	{U'} && U && X && {{\bf \Omega} }
	\arrow[from=1-5, to=1-7]
	\arrow[tail, from=1-5, to=3-5]
	\arrow["{{\bf True}}", from=1-7, to=3-7]
	\arrow[dashed, from=3-1, to=1-5]
	\arrow["f", from=3-1, to=3-3]
	\arrow[dashed, from=3-3, to=1-5]
	\arrow["\alpha"', from=3-3, to=3-5]
	\arrow["{\phi(x)}", from=3-5, to=3-7]
\end{tikzcd}
\end{small}

    \item {\bf Local character:} Analogously, if $f: U' \rightarrow U$ is an epic arrow, then from $U' \Vdash \phi(\alpha \circ f)$, we can conclude $U \Vdash \phi(x)$. 
\end{itemize}


\begin{theorem}
    If $\alpha: N \rightarrow M$ is a generalized element of causal model $M$, and $\phi(x)$ and $\psi(x)$ are formulas with a free variable $x$ of type $M$, we can conclude that
    \begin{enumerate}
        \item $N \Vdash \phi(\alpha) \wedge \psi(\alpha)$ holds if $N \Vdash \phi(\alpha)$ and $N \Vdash \psi(\alpha)$. 
        \item $N \Vdash \phi(x) \vee \psi(x)$ holds if there are morphisms $p: O \rightarrow N$ and $q: P \rightarrow N$ such that $p + q: N + O \rightarrow M$ is an epic arrow, and $N \Vdash \phi(\alpha p)$ and $O \Vdash \phi(\alpha q)$. 
        \item $N  \Vdash \phi(\alpha) \Rightarrow \psi(\alpha)$ if it holds that for any morphism $p: N \rightarrow M$, where $N \Vdash \phi(\alpha p)$, the assertion $N \Vdash \phi(\alpha p)$  also holds. 

        \item $N \Vdash \neg \phi(\alpha)$ holds if whenever the morphism $p: M \rightarrow N$ satisfies the property $N \Vdash \phi(\alpha p)$, then $N \cong {\bf 0}$. 

        \item $M \Vdash \exists \phi(x,y)$ holds if there exists an epic arrow $p: N \rightarrow M$ and generalized elements $\beta: V \rightarrow Y$ such that $M \Vdash \phi(\alpha p, \beta)$. 

        \item $M \Vdash \forall y \phi(x, y)$ holds if for every structural causal model $N$, and every arrow $p: N \rightarrow M$, and every generalized element $\beta: N \rightarrow O$, it holds that $V \Vdash \phi(\alpha p, \beta)$. 
    \end{enumerate}
\end{theorem}

{\bf Proof:} The proof follows readily from the general result on Kripke-Joyal semantics for the Mitchell-B\'enabou languages of any topos \citep{maclane1992sheaves}  The Kripke-Joyal semantics takes on a simpler form when using a Grothendieck topology on a topos, and we postpone the details to the Supplementary Materials.  \qed


\subsection{Kripke-Joyal Semantics for TCMs over Sheaves}

In the main paper, we introduced Kripke-Joyal semantics for any topos. These semantics can be specialized to a topos equipped with a Grothendieck topology, that is a site. This specialized structure captures how causal inference is woven in the fabric of the internal logic of a causal topos. Define ${\tt Sh}({\cal C, J})$ be a topos of sheaves with a specified Grothendieck topology ${\cal J}$, defined by the following diagram: 
\[ {\cal C} \xrightarrow[]{\yo} {\cal P(C)} \xrightarrow[]{a} {\tt Sh}({\cal C,J}) \cong {\cal C}\]
where we know that the Yoneda embedding $\yo$ creates a full and faithful copy of the original category ${\cal C}$. Let us define the semantics for a sheaf element $\alpha \in X(C)$, where $ X(C) = {\tt Sh}({\cal C}, J)({\cal C}(-, C), X))$. Since we know that $\{x | \phi(x) \}$ is a subsheaf, and given an arrow $f: D \rightarrow C$ of ${\cal C}$, and $\alpha \in X(C)$, then if $\alpha$ is one of the elements that satisfies the property that $\{x | \phi(x) \}$, the monotonicity property stated above implies that $\alpha \circ f \in \{ x | \phi(x) \}(D) \subseteq X(D)$. Also, the local character condition stated above implies that if $\{f_i: C_i \rightarrow C \}$ is a cover in the Grothendieck topology ${\cal J}$ such that $C_i | \Vdash \phi(\alpha \circ f_i)$ for all $i$, then $C \Vdash \phi(\alpha)$. 

With these assumptions, we can restate the Kripke-Joyal semantics for the topos category of sheaves as follows: 

\begin{enumerate}
    \item $C \Vdash \phi(\alpha) \wedge \psi(\alpha)$ if it holds that $C \Vdash \phi(\alpha)$ and $C \Vdash \psi(\alpha)$. 
    \item $C \Vdash \phi(\alpha) \vee \psi(\alpha)$ if there is a covering $\{ f_i: C_i \rightarrow C \}$ such that for each $i$, either $C_i \Vdash \phi(\alpha)$ or $C_i \Vdash \psi(\alpha)$. 
    \item $C \Vdash \phi(\alpha) \rightarrow \psi(\alpha)$ if for all $f: D \rightarrow C$, and $D \Vdash \phi(\alpha \circ f)$, it holds that $D \Vdash \psi(\alpha \circ f)$. 
    \item $C \Vdash \neg \phi(\alpha)$ holds if for all arrows $f: D \rightarrow C$ in ${\cal C}$, if $D \Vdash \phi(\alpha \circ f)$ holds, then the empty family is a cover of $D$. 
    \item $C \Vdash \exists y \ \phi(x, y)$ holds if there is a covering $\{ f_i: C_i \rightarrow C \}$ and elements $\beta_i \in Y(C_i)$ such that $C_i \Vdash \phi(\alpha \circ f_i, \beta_i)$ holds for each $i$. 
    \item Finally, for universal quantification, $C \Vdash \forall y \ \phi(x, y)$ holds if for all arrows $f: D \rightarrow C$ in the category ${\cal C}$, and all $\beta \in Y(D)$, it holds that $D \Vdash \phi(\alpha \circ f, \beta)$. 
\end{enumerate}

Summarizing this somewhat long introduction to category theory, we defined the basic notion of a category and functor, introduced natural transformations and the Yoneda Lemma, and then defined a topos. We then showed that we can define a topology on the presheafs constructed through Yoneda emebddings to give us a topos equipped with a topology. This structure gives us a general way to define causal inference  over topos categories, as we can now define subobjects over the Grothendieck sieves.  Finally, we specified the Kripke-Joyal semantics for the Mitchell-B\'enabou internal language of a topos, and we also showed that for the specific case of sheaves constructed with the $\yo$ Yoneda embedding, what the resulting semantics looked like. 

\subsection{Local Set Theory for TCMs}
\label{lst}

The Mitchell-B\'enabou language is an example of a ``local set theory" \citep{bell}. Formally, the Mitchell-B\'enabou language for the Generalized Do-Calculus is a local set theory, defined by a set of types that correspond to each structural causal model $M$ object in ${\cal C}_{TCM}$.  A {\em local set theory} \citep{bell} is defined as a language ${\cal L}$ specified by the following classes of symbols: 

\begin{enumerate}
    \item Symbols ${\bf 1} $ and $\Omega$ representing the {\em unity} type and {\em truth-value} type symbols. 

    \item A collection of symbols ${\bf A}, {\bf B}, {\bf C}, \ldots $ called {\em ground type symbols}. 

    \item A collection of symbols ${\bf f}, {\bf g}, {\bf h}, \ldots$ called {\em function} symbols. 
\end{enumerate}

To instantiate this definition for our paper, the ground types will be SCMs, each of which will be interpreted as a primitive type in Section~\ref{mbl}. We will use the topos-theoretical constructions to construct composite types. We can use an inductive procedure to recursively construct {\bf type symbols} of ${\cal L}$ as follows: 

\begin{enumerate}
    \item  Symbols ${\bf 1} $ and $\Omega$ are type symbols (the terminal object and the subobject classifier in a causal topos). 

    \item Any ground type symbol is a type symbol. For a causal topos, each SCM is a ground type symbol. 

    \item If ${\bf A}_1, \ldots, {\bf A}_n$ are type symbols, so is their product ${\bf A}_1 \times \ldots {\bf A}_n$, where for $n=0$, the type of $\prod_{i=1}^n {\bf A}_i$ is ${\bf 1}$. The product ${\bf A}_1 \times \ldots {\bf A}_n$ has the {\em product type} symbol.  These constructs allow defining an algebra of causal models. 

     \item If ${\bf A}$ is a type symbol, so is ${\bf PA}$. The type ${\bf PA}$ is called the {\em power} type. \footnote{Note that in a topos, these will be interpreted as {\em power objects}, generalizing the concept of power sets.}  We thus can give meaning to concept of a ``powerset" of a causal model, where we interpret the subobject classifier as defining the abstract semantics of a powerset for each SCM. 
\end{enumerate}

Thus, a product of SCMs will define product types. Given an SCM $M$, we can define its power type as well, which is an abstract notion of the ``power set" of a causal model (if you interpret this in the context of subobject classifiers, it means that we are defining a family of submodels). 
For each type symbol ${\bf A}$, the language ${\cal L}$ contains a set of {\em variables} $x_{\bf A}, y_{\bf A}, z_{\bf A}, \ldots$. In addition, ${\cal L}$ contains the distinguished ${\bf *}$ symbol. Each function symbol in ${\cal L}$  is assigned a {\em signature} of the form ${\bf A} \rightarrow {\bf B}$. \footnote{In a topos, these will correspond to arrows of the category.} We can define the {\em terms} of the local set theory language ${\cal L}$ recursively as follows: 

\begin{itemize}
    \item ${\bf *}$ is a term of type ${\bf 1}$. 

    \item for each type symbol ${\bf A}$, variables $x_{\bf A}, y_{\bf A}, \ldots$ are terms of type ${\bf A}$. 

    \item if ${\bf f}$ is a function symbol with signature ${\bf A} \rightarrow {\bf B}$, and $\tau$ is a term of type ${\bf A}$, then ${\bf f}(\tau)$ is a term of type ${\bf B}$. 

    \item If $\tau_1, \ldots, \tau_n$ are terms of types ${\bf A}_1, \ldots, {\bf A}_n$, then $\langle \tau_1, \ldots \tau_n \rangle$ is a term of type ${\bf A}_1 \times \ldots {\bf A}_n$, where if $n=0$, then $\langle \tau_1, \ldots \tau_n \rangle$ is of type ${\bf *}$. 

    \item If $\tau$ is a term of type ${\bf A}_1 \times {\bf A}_n$, then for $1 \leq i \leq n$, $(\tau)_i$ is a term of type ${\bf A}_i$. 

    \item if $\alpha$ is a term of type $\Omega$, and $x_{\bf A}$ is a variable of type ${\bf A}$, then $\{x_{\bf A} : \alpha \}$ is a term of type ${\bf PA}$.  

    \item if $\sigma, \tau$ are terms of the same type, $\sigma = \tau$ is a term of type $\Omega$. 

    \item if $\sigma, \tau$ are terms of the types ${\bf A}, {\bf PA}$, respectively, then $\sigma \in \tau$ is a term of type ${\bf \Omega}$. 
\end{itemize}

A term of type ${\bf \Omega}$ is called a {\em formula}. The language ${\cal L}$ does not yet have defined any logical operations, because in a typed language, logical operations can be defined in terms of the types, as illustrated below. 

\begin{itemize}
    \item $\alpha \Leftrightarrow \beta$ is interpreted as $\alpha = \beta$. 

    \item {\bf true} is interpreted as ${\bf *} = {\bf *}$. 

    \item $\alpha \wedge \beta$ is interpreted as $\langle \alpha, \beta \rangle = \langle {\bf true}, {\bf false} \rangle$. 

    \item $\alpha \Rightarrow \beta$ is interpreted as $(\alpha \wedge \beta) \Leftrightarrow \alpha$

    \item $\forall x \ \alpha$ is interpreted as $\{x : \alpha\} = \{x : {\bf true} \}$

    \item ${\bf false}$ is interpreted as $\forall \omega \ \omega$. 

    \item $\neg \alpha$ is interpreted as $\alpha \Rightarrow {\bf false}$.

    \item $\alpha \vee \beta$ is interpreted as $\forall \omega \ [(\alpha \Rightarrow \omega \wedge \beta \Rightarrow \omega) \Rightarrow \omega]$

    \item $\exists x \ \alpha$ is interpreted as $\forall \omega [ \forall x (\alpha \Rightarrow \omega) \Rightarrow \omega ]$

\end{itemize}

Finally, we have to specify the inference rules, which are given in the form of {\em sequents}. We will just sketch out a few, and the rest can be seen in \citep{bell}. A sequent is a formula $\Gamma: \alpha$ where $\alpha$ is a formula, and $\Gamma$ is a possibly empty finite set of formulae. The basic axioms include $\alpha: \alpha$ (tautology), $:x_1 = {\bf *}$ (unity), a rule for forming projections of products, a rule for equality, and another for comprehension. Finally, the inference rules are given in the form: 

\begin{itemize}
    \item {\em Thinning:}
    \[
  \begin{prooftree}
    \hypo{\Gamma : \alpha}
    \infer1{\beta, \Gamma: \alpha}
  \end{prooftree}
\]
\item {\em Cut}: 

\[
  \begin{prooftree}
    \hypo{\Gamma : \alpha, \  \ \alpha, \Gamma: \beta}
    \infer1{\Gamma: \beta}
  \end{prooftree}
\]

\item {\em Equivalence}: 

\[
  \begin{prooftree}
    \hypo{\alpha, \Gamma : \beta \ \ \beta, \Gamma: \alpha}
    \infer1{\Gamma: \alpha \Leftrightarrow \beta}
  \end{prooftree} 
\]

\end{itemize}

A full list of inference rules with examples of proofs is given in \citep{bell}. Now that we have the elements of a local set theory defined as shown above, we need to connect its definitions with a causal topos. That is the topic of the next section.

\section{Proofs of the Theorems in Main Paper}
\label{proofs} 

We now include more detailed proofs of the key theorems in the main paper, building on the background sections above. 

\setcounter{theorem}{0}
\begin{theorem}
    The  category ${\cal C}_{TCM}$ forms a topos. 
\end{theorem}

{\bf Proof:} To show this result, we need to check that the conditions for being a topos are satisfied. The proof can be constructed in several ways, depending on which sets of conditions are used. We already showed in the main paper that the category ${\cal C}_{TCM}$ has a subobject classifier (see Figure\ref{subobj-classifier}). So, we focus here primarily on the other conditions. Note that the terminal object is simply the identity function ${\bf id}_0: \{0 \} \rightarrow \{0 \}$. Now, it remains to show that ${\cal C}_{TCM}$ has pullbacks and exponential objects. 

{\bf Pullbacks in ${\cal C}_{TCM}$}: Consider the cube shown in Figure~\ref{pullback-gdc}. Here, $f$, $g$, and $h$ can be interpreted as three TCMs, each mapping some exogenous variables to some endogenous variables. The arrows $i, j$ ensure that the bottom face of the cube is a commutative diagram, and the arrows $p, q$ ensures the right face of the cube is a commutative diagram. The arrow from $P$ to $Q$ exists because looking at the front face of the cube, $Q$ is the pullback of $i$ and $q$, which must exist because we are in the category of {\bf Sets}, which has all pullbacks. Similarly, the back face of the cube is a pullback of $j$ and $p$, which is again a pullback in {\bf Sets}. Summarizing, $\langle u, v \rangle$ and $\langle m, n \rangle$ are the pullbacks of $\langle i, j \rangle$ and $\langle p, q \rangle$. 

{\bf Exponential objects in ${\cal C}_{TCM}$}: Now it only remains to check that the category has exponential objects. Let $f: U \rightarrow V$ and $g: U' \rightarrow V'$ be two functions induced by TCM models $M$ and $N$. Then, we need to define the meaning of $g^f$ in ${\cal C}_{TCM}$, which we can define as $g^f: X \rightarrow Y$, where $Y = V'^{V}$, which must exist since {\bf Sets} is a Cartesian closed category that has exponential objects (i.e., $Y$ is simply the set of all functions from $V$ to $V'$). Also, $X$  is the set of all arrows in ${\cal C}_{TCM}$ from SCM $M$ to SCM $N$, which is the pair of functions $\langle h, k \rangle$ in the commutative diagram shown below. This finally proves that ${\cal C}_{TCM}$ is a topos. 
\begin{center}
\begin{tikzcd}
	U && {U'} \\
	\\
	V && {V'}
	\arrow["h"', from=1-1, to=1-3]
	\arrow["f", from=1-1, to=3-1]
	\arrow["g"', from=1-3, to=3-3]
	\arrow["k", from=3-1, to=3-3]
\end{tikzcd}
\end{center} \qed 
\begin{figure}
    \centering
    \includegraphics[width=0.3\linewidth]{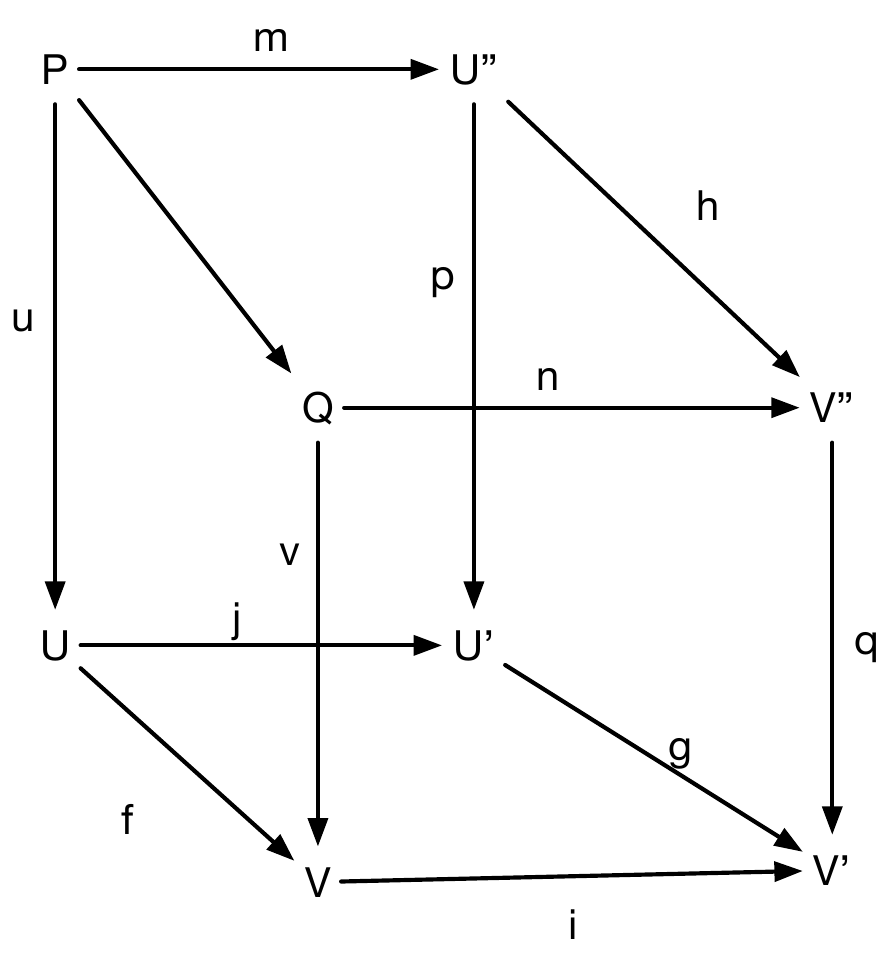}
    \caption{Figure showing that ${\cal C}_{TCM}$ has pullbacks.}
    \label{pullback-gdc2}
\end{figure}
\begin{theorem}
    The functor category of presheaves over Bayesian Networks modeled as functors between CDU symmetric monoidal categories forms a topos. 
\end{theorem}
\begin{theorem}
    The functor category of presheaves over Markov categories forms a topos. 
\end{theorem}
\begin{theorem}
    The functor category over causal models represented as simplicial sets forms a topos. 
\end{theorem}
\begin{figure}
    \centering
    \includegraphics[width=0.5\linewidth]{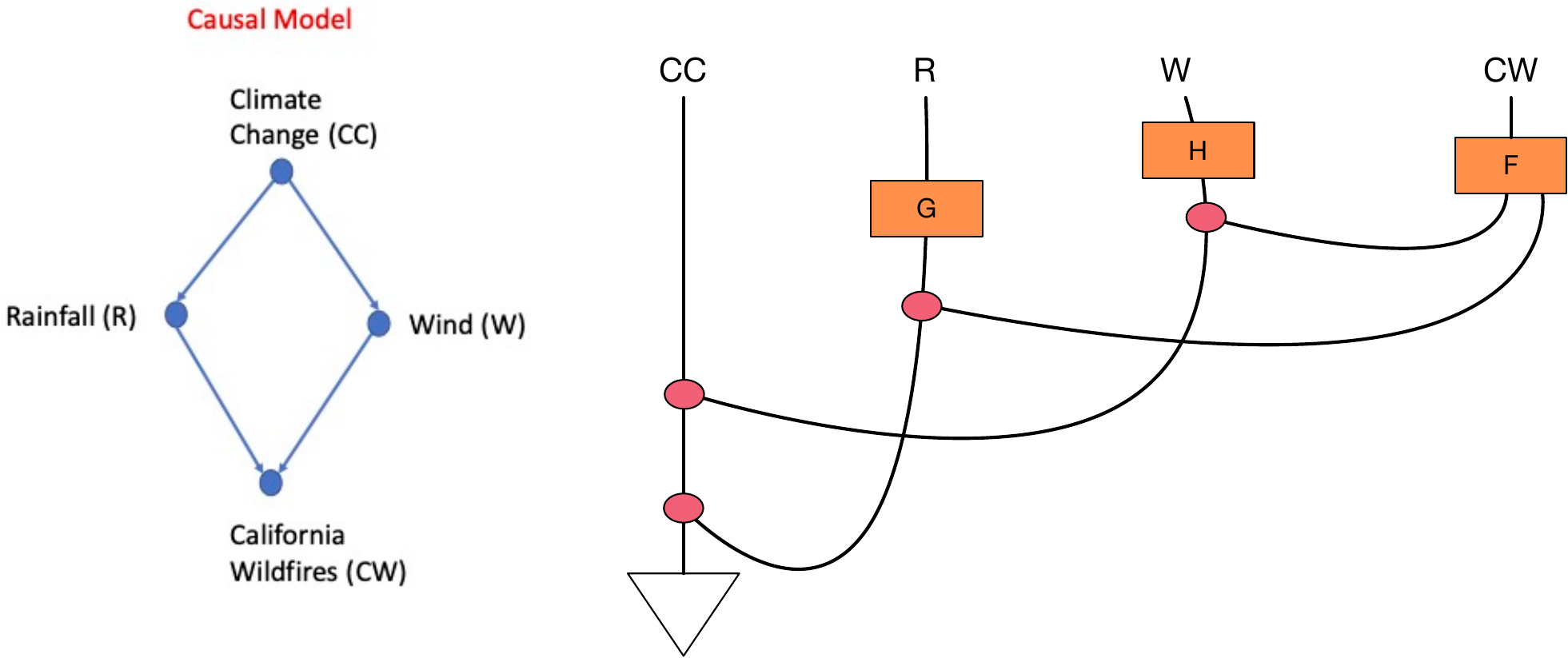}
    \caption{String Diagram of a simple climate change model for California.}
    \label{string-diagram}
\end{figure}
{\bf Proof:} Before giving the proof of these theorems, let us build some intuition about these categories. Each object in the topos category $C_{TCM}$ is now a string diagram causal model, as shown in Figure~\ref{string-diagram}.  We can interpret causal interventions in such string diagrams as ``string diagram surgery", more formally an endofunctor on the functor category ${\bf FiniteStoch}^{{\cal C}_{\mbox{CDU}}}$. Intuitively, each object in this category is a Bayes network, which is represented as a functor from an affine CDU (or Markov) category to the symmetric monoidal category {\bf FinStoch} of finite stochastic processes. \citet{string-diagram-surgery} define {\bf FinStoch} in detail, but essentially, it has natural numbers $n > 0$ as objects, and morphisms from $m \rightarrow n$ are column stochastic matrices. We can construct topoi from affine CDU and Markov categories using the Yoneda embedding.  Intuitively, the string diagram represents the induced function of a TCM as a network of nodes and computing units (see Figure~\ref{string-diagram}), where the downward directed triangle indicates an exogenous variable ($CC$) and the other variables are endogenous variables. Applying a do(X=x) to any subset $X$ of $V = \mbox{R, W, CW} $ produces  submodel that can be viewed as a subobject of the string diagram shown in Figure~\ref{string-diagram}. We can specifically define a subobject classifier over string diagrams as well

We can prove all three theorems using standard  results in the theory of sheaves \citep{maclane1992sheaves}, but let us give some intuition in a simple setting. In the category {\bf {Sets}}, we know that every object (i.e., a set) $X$ can be expressed as a coproduct (i.e., disjoint union)  of its elements $X \simeq \sqcup_{x \in X} \{ x \}$, where $x \in X$. Note that we can view each element $x \in X$ as a morphism $x: \{ * \} \rightarrow X$ from the one-point set to $X$. The categorical generalization of this result is called the {\em {density theorem}} in the theory of sheaves. Let $F: {\cal D} \rightarrow {\cal C}$ be a functor from category ${\cal D}$ to ${\cal C}$. The {\bf {comma category}} $F \downarrow {\cal C}$ is one whose objects are pairs $(D, f)$, where $D \in {\cal D}$ is an object of ${\cal D}$ and $f \in$ {\bf {Hom}}$_{\cal C}(F(D), C)$, where $C$ is an object of ${\cal C}$. Morphisms in the comma category $F \downarrow {\cal C}$ from $(D, f)$ to $(D', f')$, where $g: D \rightarrow D'$, such that $f' \circ F(g) = f$. We can depict this structure through the following commutative diagram: 
\begin{center} 
\begin{tikzcd}[column sep=small]
& F(D) \arrow{dl}[near start]{F(g)} \arrow{dr}{f} & \\
  F(D')\arrow{rr}{f'}&                         & C
\end{tikzcd}
\end{center} 
Let {\cal D} be a small category, {\cal C} be an arbitrary category, and $F: {\cal D} \rightarrow {\cal D}$ be a functor. The functor $F$ is {\bf {dense}} if for all objects $C$ of ${\cal C}$, the natural transformation 
\[ \psi^C_F: F \circ U \rightarrow \Delta_C, \ \ (\psi^C_F)_{({\cal D}, f)} = f\]
is universal in the sense that it induces an isomorphism $\mbox{Colimit}_{F \downarrow C} F \circ U \simeq C$. Here, $U: F \downarrow C \rightarrow {\cal D}$ is the projection functor from the comma category $F \downarrow {\cal C}$, defined by $U(D, f) = D$. A fundamental consequence of the category of elements is that every object in the functor category of presheaves, namely contravariant functors from a category into the category of sets, is the colimit of a diagram of representable objects, via the Yoneda lemma. Notice this is a generalized form of the density notion from the category {\bf {Sets}}. In the functor category of presheaves {\bf {Set}}$^{{\cal C}^{op}}$, every object $P$ is the colimit of a diagram of representable objects, in a canonical way. 
To construct the canonical diagram, we can use the standard result that the presheaf $P$ is isomorphic to the colimit of the diagram below: 
\[ P \cong \mbox{Colim} \{ \int P \xrightarrow[]{\pi_P} {\cal C} \xrightarrow[]{\yo} \hat{C} \}\]
where $\hat{C} = {\bf Sets}^{C^{op}}$, a complete and cocomplete category of presheaves defined by the Yoneda embedding $\yo(x) = {\cal C}(-,x)$, and $\pi_P$ is the projection from the category of elements to the category ${\cal C}$.  Finally, another standard result from \citep{maclane1992sheaves} is that the Yoneda embedding $\yo$ produces a topos, namely the TCM category $\hat{C}$ produced by the Yoneda embedding is Cartesian closed, has pullbacks, and has a subobject classifier, thereby forming a topos. Combining these results, they imply that a TCM category constructed from $\yo$ Yoneda embeddings of affine CDU categories,  Markov categories, or simplicial sets forms a topos. \qed 

\end{document}